# Group Communication Analysis: A Computational Linguistics Approach for Detecting Sociocognitive Roles in Multi-Party Interactions

Nia M. M. Dowell[1], Tristian M. Nixon, & Arthur Graesser[2]

*University of Michigan, Ann Arbor, Michigan, USA*
*University of Memphis, Memphis, Tennessee, USA*

## Abstract

Roles are one of the most important concepts in understanding human sociocognitive behavior. During group interactions, members take on different roles within the discussion. Roles have distinct patterns of behavioral engagement (i.e., active or passive, leading or following), contribution characteristics (i.e., providing new information or echoing given material), and social orientation (i.e., individual or group). Different combinations of these roles can produce characteristically different group outcomes, being either less or more productive towards collective goals. In online collaborative learning environments, this can lead to better or worse learning outcomes for the individual participants. In this study, we propose and validate a novel approach for detecting emergent roles from the participants' contributions and patterns of interaction. Specifically, we developed a group communication analysis (GCA) by combining automated computational linguistic techniques with analyses of the sequential interactions of online group communication. The GCA was applied to three large collaborative interaction datasets (participant $N = 2{,}429$; group $N = 3{,}598$). Cluster analyses and linear mixed-effects modeling were used to assess the validity of the GCA approach and the influence of learner roles on student and group performance. The results indicate that participants' patterns in linguistic coordination and cohesion are representative of the roles that individuals play in collaborative discussions. More broadly, GCA provides a framework for researchers to explore the micro intra- and inter-personal patterns associated with the participants' roles and the sociocognitive processes related to successful collaboration.

*Keywords: Group communication analysis, social roles, group interaction, computational linguistics*

## Introduction

Roles are one of the most important concepts in understanding human sociocognitive behavior. When individuals engage in everyday interactions, they act in ways that are both enabled and constrained by social structure: the social context, history, structures of interaction, and the attributes which individuals bring to the interaction (Gleave, Welser, Lento, & Smith, 2009; Hare, 1994; Sapru & Bourlard, 2015). In this context, social roles provide a valuable window into the underlying sociocognitive structure of group interaction, and one that researchers can use to differentiate individuals and explain the consequences of an individual's and overall group's behavior (Gleave et al., 2009; Mudrack & Farrell, 1995).

The concept of social roles has garnered significant interdisciplinary attention across several areas including education, social computing, and social and organizational psychology. This has produced a burgeoning literature on social roles in a variety of domains, including: teams (Driskell, Driskell, Burke, & Salas, 2017), workplace meetings (Sapru & Bourlard, 2015), and collaborative interactions (Strijbos & De Laat, 2010). Roles have been defined more strictly as stated functions and/or responsibilities that guide individual behavior, and behavioral patterns exemplified by individuals in social contexts (Chiu, 2000; Hare, 1994; Volet, Vauras, Salo, & Khosa, 2017). This definition is reflected in the two prominent perspectives on roles that appear in the sociological and psychological literature. The first emphasizes the behaviors associated with a specific appointment in a group or in an organization. The most obvious examples falling under this class are formal appointments, like employment positions, political offices, military ranks, academic degrees, and other formal titles. However, this category also includes roles in more *ad hoc* social situations, such as those that are explicitly teacher-assigned for the purposes of some exercise, or implicitly embodied through pre-scripted interactions. In this context, the role is a position to which a person is assigned and then performs the behavior associated with that position (Salazar, 1996). The second perspective, by contrast, considers roles as a product of a specific interaction context, consisting of patterns in sociocognitive behaviors enacted by people (Gleave et al., 2009). These roles are emergent in that they develop naturally out of the interpersonal interaction without any prior instruction or assignment, and are defined (characterized) by their behavioral proximity (similarities and differences) to other interactional partners.

Several studies have emphasized the importance of roles in group interactions (Dillenbourg, 1999; Hoadley, 2010; Jahnke, 2010; Marcos-Garcia, Martinez-Mones, & Dimitriadis, 2015; Sarmiento & Shumar, 2010; Smith Risser & Bottoms, 2014; Spada, 2010; Stahl, Law, Cress, & Ludvigsen, 2014; Strijbos & De Laat, 2010; Volet et al., 2017). Recent work on scripted or assigned roles shows that the assignment of specific roles facilitates collaborative awareness (Strijbos, Martens, Jochems, & Broers, 2004), team discourse and performance (Gervits, Eberhard, & Scheutz, 2016; Xie, Yu, & Bradshaw, 2014), and the depth of knowledge co-construction by the group (De Wever, Keer, Schellens, & Valcke, 2010; Gu, Shao, Guo, & Lim, 2015). While assigning roles to group members may produce beneficial outcomes, there are concerns. First, the occurrence of potentially dysfunctional group roles is generally neglected in the literature, with researchers choosing to focus on those roles that are potentially most productive for the group as opposed to the roles that actually exist (Lehmann-Willenbrock, Beck, & Kauffeld, 2016). Considering the prevalence of dysfunctional group behavior (e.g., the "bad apple phenomenon"; Felps, Mitchell, & Byington, 2006), deepening our understanding of such negative roles and their influence is crucial. This leads to the second concern, regarding what is captured in role assignment research. Simply because someone is assigned a role, does not mean they will not deviate from said role. Are we, then, exploring roles-as-intended or roles-as-enacted (Hoadley, 2010)? Finally, by attempting to restrict an individual to a single role one inhibits role and group flexibility, and the potential advantages of this. It also disregards the dynamic and interactive way in which roles are created, negotiated, and evolve among group members during social interaction (Hoadley, 2010; Lehmann-Willenbrock et al., 2016; Salazar, 1996).

Researchers have attempted to detect the emergence of roles during online group interactions (Stahl et al., 2014). The majority of these efforts have relied on predefined content analysis coding schemes and complex taxonomies to determine what roles each individual occupies within the group (e.g., Arvaja & Hämäläinen, 2008; Volet et al., 2017). For instance, Strijbos and De Laat (2010) provided a valuable conceptual framework of roles in group interactions. Their framework distinguishes eight roles. Four of the roles are reserved for large group interactions: Pillar, Generator, Hanger-on and Lurker. However, the remaining four are particularly relevant to small group interactions: Captain, Over-rider, Free-rider, and Ghost. The roles are differentiated along two dimensions that crosses orientation (individual, group) and effort (low, high). The Strijbos and De Laat (2010) framework helped guide some of our initial conceptualizations of the processes involved in participant roles. However, as will be shown, we adopted an automated methodological approach that afforded several new dimensions of interaction. Particularly, while extensive knowledge has been gleaned from manual content analyses of emergent social roles, several researchers have pointed to its inherent limitations. These practices tend to obscure the sequential structure, semantic references within group discussion, and situated methods of interaction through which roles emerge (Çakır, Zemel, & Stahl, 2009; Strijbos, Martens, Prins, & Jochems, 2006; Suthers, Dwyer, Medina, & Vatrapu, 2010). Moreover, manual coding methods are no longer a viable option with the increasing scale of online group interaction data (Daradoumis, Martínez-Monés, & Xhafa, 2006).

The availability of such data represents a golden opportunity to make advances in understanding social roles and role ecologies (Gleave et al., 2009). However, automatic approaches for detecting emergent social roles is still relatively scarce in the field of collaborative interactions. The attempts that have been made typically rely on social network analysis (SNA) (e.g., Capuano, Mangione, Mazzoni, Miranda, & Orciuoli, 2014; Marcos-Garcia et al., 2015; Stuetzer, Koehler, Carley, & Thiem, 2013). In this context, social roles are characterized in terms of behavioral regularities and network attributes, wherein consistent behaviors resulting in persistent or recurrent interactions between individuals in a social group are potential signals of a meaningful social role (Gleave et al., 2009). One of the advantages of quantitative methods like SNA is that alleviates the human time requirement and the attendant subjectivity issues inherent in manual content analyses. However, these strictly structural measures have been criticized for being only surface level, because they do not capture the deeper level interpersonal sociocognitive and semantic information found in the discourse interaction (Strijbos & Weinberger, 2010). Automated natural language processing techniques could provide a productive path towards automated role detection by addressing some of these limitations. Specifically, roles emerge and are sustained through interaction (Hare, 1994; Salazar, 1996), and communication is the basis of any interaction. Indeed, language and communication has proven quite useful in other explorations of group interaction phenomena (Cade, Dowell, Graesser, Tausczik, & Pennebaker, 2014; Cai et al., 2017; Dowell, Brooks, Kovanović, Joksimović, & Gašević, 2017; Dowell et al., 2017; Dowell, Cade, Tausczik, Pennebaker, & Graesser, 2014; Dowell, & Graesser, 2015; Dowell et al., 2015; Graesser, Dowell, & Clewley, 2017; Ho et al., 2016; Joksimović et al., in press, 2015). As such, language provides a powerful and measureable behavioral signal that can be used to capture the semantic and sociocognitive interaction patterns that characterize emergent roles, as well as to study their influence on the outcomes of group interactions.

## Overview of Present Research

The current research has two main objectives. The first is to propose an automated methodology, *Group Communication Analysis* (GCA), for detecting emergent roles in group interactions. The GCA combines computational linguistic techniques with sequential interaction analyses of group communication. The GCA captures theoretically relevant sociocognitive processes that can be used to characterize the ***social roles*** individuals occupy in group interactions. Tracking the communication dynamics during ongoing group interactions can reveal important patterns about how individual and group processes emerge and unfold over time. The second goal of this research is to explore how the individual-level roles and overall group compositions influence both student and group performance during collaborative interactions. The concepts, methods, and ideas presented in this research are at the intersection of collaborative learning, discourse processes, data mining, and learning analytics. This interdisciplinary research approach will hopefully provide insights and help redefine the nature of roles in group interaction. Specifically, the current research includes analyses of two large, collaborative learning datasets (Traditional CSCL learner $N = 854$, group $N = 184$; SMOC learner $N = 1,713$, group $N = 3,297$), and one collaborative problem-solving dataset (Land Science learner $N = 38$; group $N = 630$) to address the research questions outlined below. While this investigation takes place in the context of collaborative learning, the methodology is flexible and could be applied in any computer-mediated social interaction space that involves linguistic interactions between the participants.

### Research Questions

1. Can individual roles be identified through patterns of communication and participation during collaborative interactions of some specific type or context?

2a. Do the patterns, if any, observed from research question 1 generalize meaningfully to other collaborative interactions of the same type or context?

2b. Do the patterns, if any, observed from research question 1 generalize meaningfully to other collaborative interactions of different types or contexts?

3a. How does an individual's role influence individual and group performance?

3b. How does group role diversity and composition influence individual and group performance?

The subsequent sections of the paper are organized as follows. First, we provide the theoretical foundation for the GCA measures, followed by a detailed technical description of the construction of the measures. We then move into the methodological features of the current investigation, followed by the details of the cluster analysis that was used to identify specific individual roles in the communication patterns during collaborative interactions. Next, we discuss the linear mixed-effects modeling used to assess the validity of the GCA approach and the influence of roles on individual and group performance. We conclude the paper with a detailed discussion of the results in the context of theory, as well as a general discussion of the theoretical, methodological, and practical implications for group interaction research.

## Group Communication Analysis (GCA)

### Theoretical Motivation for the GCA Measures

Social and cognitive processes are the fabric of collaborative learning. The ultimate goal of collaborative learning is the co-constructed knowledge that results from the sharing of information in groups during collaborative tasks (Alavi & Dufner, 2004; Dillenbourg & Fischer, 2007). Learning as a social process is supported by several theoretical perspectives including the social cognitive theory (Bandura, 1994), social-constructivist framework (Doise, 1990), socio-cultural framework (Vygotsky, 1978), group cognition models (Stahl, 2005), shared cognition theory (Lave & Wenger, 1991), and connectivism (Siemens, 2005). Research on the sociocognitive aspects of computer-supported collaborative learning (CSCL) have noted some of the important mechanisms (e.g., social presence, explanation, negotiation, monitoring, grounding, and regulating) and processes (e.g., convergence, knowledge co-construction, meaning-making) that facilitate successful outcomes (Dillenbourg, Järvelä, & Fischer, 2009).

The Group Communication Analysis framework incorporates definitions and theoretical constructs that are based on research and best practices from several areas where group interaction and collaborative skills have been assessed. These areas include computer-supported cooperative work, team discourse analysis, knowledge sharing, individual problem solving, organizational psychology, and assessment in work contexts (e.g., military teams, corporate leadership). The framework further incorporates information from existing assessments that can inform the investigation of social roles, including the PISA 2015 CPS Assessment. Despite differences in orientation between the disciplines where these frameworks have originated, the conversational behaviors that have been identified as valuable are quite similar. The following sections review the theoretical perspectives and sociocognitive processes that were the foundation the GCA framework and resulting metrics (i.e., Participation, Internal Cohesion, Responsivity, Social Impact, Newness and Communication Density). In the presentation of the theoretical principles and sociocognitive processes supporting the GCA metrics, empirical findings are presented whenever possible as illustration and initial support. Table 1 provides a summary of the alignment of the GCA dimensions with associated theoretical and empirical support.

**Participation**. Participation is obviously a minimum requirement for collaborative interaction. It signifies a willingness and readiness of participants to externalize and share information and thoughts (Care et al., 2016; Hesse et al., 2015). Previous research has confirmed that participation, measured as interaction with peers and teachers, has a beneficial influence on perceived and actual learning, retention rates, learner satisfaction, social capital, and reflection (Hew, Cheung, & Ng, 2010; see Hrastinski, 2008 for a review). Within collaborative groups, individual students who withdraw their participation from group discussion or only participate minimally can undermine learning, either because of lost opportunities for collaboration or by provoking whole group disengagement (Van den Bossche, Gijselaers, Segers, & Kirschner, 2006). In CSCL research, typical measures of student participation include the number of a student's contributions (Lipponen, Rahikainen, Lallimo, & Hakkarainen, 2003), length of posts in online environments (Guzdial & Turns, 2000), or whether contributions are more social (i.e., off-task) rather than focused on content ideas (Stahl, 2000). More recently, Wise and colleagues (2012) have argued that a more complete conception of participation in online discussions requires attention not only to participants' overt activity in producing contributions, but also to the less public activity of interacting with the contributions of others, which they have termed "online listening behavior" (Wise, Speer, Marbouti, & Hsiao, 2012). Taken together, this research highlights how individual participants may vary in the amount, type, and quality of participation within a group. Therefore, participation is an important metric to characterize the social roles participants occupy during group interactions. In the current research, ***participation*** is conceptualized as a necessary, but not sufficient, sociocognitive metric for characterizing the participants' social roles.

   **Internal cohesion, responsiveness, & social impact**. Simply placing participants in groups does not guarantee collaboration or learning (Kreijns, Kirschner, & Jochems, 2003). For collaboration to be effective, participants must participate in shared knowledge construction, have the ability to coordinate different perspectives, commit to joint goals, and evaluate their collective activities together (Akkerman et al., 2007; Beers, Boshuizen, Kirschner, & Gijselaers, 2007; Blumenfeld, Kempler, & Krajcik, 2006; Fiore & Schooler, 2004; Kirschner, Paas, & Kirschner, 2009; Roschelle & Teasley, 1995). This raises an important question that has been reoccurring theme in the CSCL literature: ***What makes collaborative discourse productive for learning?*** (Stahl & Rosé, 2013). This question has been studied with a related focus and comparable results across several CSCL sub-communities.

**Table 1** Alignment of GCA Dimensions with Theoretical and Empirical Support

| GCA Dimensions | Psychological & Discursive Processes | Description/ Example Behavioral Makers | Relevant Theoretical Frameworks & Constructs | Empirical Evidence/ Theoretical Support |
|---|---|---|---|---|
| Participation | Engagement | General level of an individual's participation, irrespective of the semantic content of this participation | Activity theory; Social presence; Socio-constructivist | Hesse et al., 2015; Hrastinski, 2008; Hew, Cheung, & Ng, 2010 |
| Internal Cohesion | Monitoring and reflecting | The tendency of an individual to consistency or novelty in their own contributions. | Self-regulation and metacognitive processes | Chan, 2012; Zimmerman, 2001; Barron, 2000; OECD, 2013 |
| Responsivity | Uptake and Transactivity | The tendency of an individual to respond, or not, to the contributions of their collaborative peers | Meaning-making; co-regulation; Interactive alignment; Social coordination; Knowledge building; Common Ground; Knowledge convergence | Berkowitz & Gibbs, 1983; Teasley, 1997; Hesse et al., 2014; Suthers, 2006; Volet et al. 2009 |
| Social Impact | Productive or popular communication | The tendency of a participant to evoke responses, or not, from their collaborative peers | Social coordination; Knowledge building; Common Ground; co-construction | Volet et al. 2009; Hesse et al., 2014; Suthers, 2006 |
| Newness | Novelty of information shared | The tendency to provide new information or to echo previously stated information, irrespective of the originator of the information. | Monitoring; Information sharing | Chi, 2009; Hesse et al., 2014; Mesmer-Magnus & Dechurch, 2009; Kirschner, Beers, Boshuizen, & Gijselaers, 2008 |
| Communication Density | Concise communication | The extent to which participants convey information in a concise manner | Common ground; Effective communication | Gorman et al. 2003; 2004 |

Collaborative knowledge construction is understood as an unequivocally interpersonal and contextual phenomenon, but the role of an individual interacting with themselves should also be taken into account (Stahl, 2002). Successful collaboration requires that each individual *monitor* and *reflect* on their own knowledge and contributions to the group (Barron, 2000; OECD, 2013). This points to the importance of self-regulation in collaborative interactions (Chan, 2012; Zimmerman, 2001). Self-regulation is described as an active, constructive process in which participants set goals, and monitor and evaluate their cognition, affects, and behavior (Azevedo, Winters, & Moos, 2004; Pintrich, 2000; Winne, 2013). During collaborative interactions, this is necessary for individuals to appropriately build on and integrate their own views with those of the group (Kreijns et al., 2003; OECD, 2013). The process of individuals engaging in self-monitoring and reflection may be reflected in their ***internal cohesion***. That is, a participant's current and previous contributions should be, to some extent, semantically related to each other, which can indicate the integration and evolution of their thoughts through monitoring and reflecting (i.e., self-regulation). However, overly high levels of internal cohesion might also suggest that a participant is not building on or evolving their thoughts, but rather are simply reiterating the same static view. Conversely, very low levels of internal cohesion might indicate that a participant has no consistent perspective on offer to the group, is simply echoing the views of others, or is only engaging at a nominal or surface level. Therefore, we should expect productive roles to exhibit a moderate degree of internal cohesion.

Participants must also monitor and build on the perspectives of their collaborative partners to achieve and maintain a shared understanding of the task and its solutions (Dillenbourg & Traum, 2006; Graesser et al., in press; Hmelo-Silver & Barrows, 2008; OECD, 2013; Stahl & Rosé, 2013). In the CSCL literature this shared understanding has been referred to as knowledge convergence, or common ground (Clark, 1996; Clark & Brennan, 1991; Fiore & Schooler, 2004; Roschelle, 1992). It is achieved through communication and interaction, such as building a shared representation of the meaning of the goal, coordinating efforts, understanding the abilities and viewpoints of group members, and mutual monitoring of progress towards the solution. These activities are supported in several collaborative learning perspectives (e.g., cognitive elaboration, Chi, 2009; socio-cognitive conflict, Doise, 1990; Piaget, 1993; co-construction, Hatano, 1993; Van Boxtel, 2004), each of which stress different mechanisms to facilitate learning during group interactions (giving, receiving and using explanations, resolving conflicts, co-construction). However, all these perspectives are in alignment on the idea that it is the participants' elaborations on one another's contributions that support learning.

These social processes of awareness, monitoring, and regulatory processes all fall under the shared umbrella of co-regulation. Volet, Summers, and Thurman (2009) proposed *co-regulation* as an extension of self-regulation to the group or collaborative context, wherein co-regulation is described as individuals working together as multiple self-regulating agents, all socially monitoring and regulating each other's learning. In a class-room study of collaborative learning using hypermedia, Azevedo et al. (2004) demonstrated that collaborative outcomes were related to the use of regulatory behaviors. In this process, the action of one student does not become a part of the group's common activity until other collaborative partners react to it. If other group members do not react to a student's contribution, this suggests the contribution was not seen as valuable by the other group members and would be an 'ignored co-regulation attempt' (Molenaar, Chiu, Sleegers, & Boxtel, 2011). Therefore, the concepts of *transactivity* and *uptake* (Table 1) in the CSCL literature are important in this context of co-regulation and active learning, in the sense that a student takes up another student's contribution and continues it (Berkowitz & Gibbs, 1983; Suthers, 2006; Teasley, 1997). Students can engage in higher or lower degrees of co-regulation through monitoring and coordinating. These processes will be represented in their discourse.

Monitoring and regulatory processes are, hopefully, externalized during communication with other group members. We can capture the degree to which an individual is monitoring and incorporating the information provided by their peers by examining the semantic relatedness between the individual's current contribution and the previous contributions of their collaborative partners. This measure is called ***responsivivity*** in the current research. For example, if an individual's contributions are, on average, only minimaly related to those of their peers, than we would say this individual has low responsivity. Similarly, we can capture the extent to which a participant's contributions are seen as meaningful, or worthy of further discussion (i.e. uptake), by their peers by measuring the semantic relatedness between the participant's current contribution and those that follow from their collaborative partners. This measure is called ***social impact*** in the current research. Participants have high social impact to the extent that their contributions are often semantically related to the subsequent contributions from the other collaborative group members.

The collaborative learning literature highlights the value of students clearly articulating arguments and ideas, as well as elaborating on, and making connections between contributions.

For instance, Rosé and colleagues' work has concentrated explicitly on properties like transactivity (Gweon, Jain, McDonough, Raj, & Rosé, 2013; Joshi & Rosé, 2007; Rosé et al., 2008), as well as the social aspects and conversational characteristics that facilitate the recognition of transactivity (Howley et al., 2011; Howley, Mayfield, & Rosé, 2013; Howley, Mayfield, Rosé, & Strijbos, 2013; Wen, Yang, & Rose, 2014). Their research adopts a sociocognitive view (Howley, Mayfield, Rosé, et al., 2013) that emphasizes the significance of publically articulating ideas and encouraging participants to listen carefully to and build on one another's ideas. Participants engaging in this type of activity have the chance to notice discrepancies between their own mental model and those of other members of the group. The discussion provides opportunities to engage in productive cognitive conflict and knowledge construction (Howley, Mayfield, Rosé, et al., 2013). Additionally, participants benefit socially and personally from the opportunity to take ownership over ideas and position themselves as valuable sources of knowledge within the collaborative group (Howley & Mayfield, 2011).

**Newness and communication density**. For collaboration to be successful, participants must engage in effective information sharing. Indeed, one of the primary advantages of collaborative interactions and teams is that they provide the opportunity to expand the pool of available information, thereby enabling groups to reach higher quality solutions than could be reached by any one individual (Hesse et al., 2015; Kirschner, Beers, Boshuizen, & Gijselaers, 2008; Mesmer-Magnus & Dechurch, 2009). However, despite the intuitive importance of effective information sharing, a consistent finding from research is that groups predominantly discuss information that is *shared* (known to all participants) at the expense of information that is *unshared* (known to a single member) (Stasser & Titus, 1985; see Wittenbaum & Stasser, 1996 for a review). This finding has been called *bias information sharing* or *bias information pooling* in the Collective Information-Sharing Paradigm. It shares some similarities with the groupthink phenomena (Janis, 1983), which is the tendency for groups to drive for consensus that overrides critical appraisal and the consideration of alternatives. The collective preference for redundant information can detrimentally affect the quality of the group interactions (Hesse et al., 2015) and decisions made within the group (Wittenbaum, Hollingshead, & Botero, 2004). However, collaborative interactions benefit when the members engage in the constructive discourse of inferring and sharing new information and integrating new information with existing prior knowledge during the interaction (Chi, 2009; Chi & Menekse, 2015).

The distinction between given (old) information versus new information in discourse is a foundational distinction in theories of discourse processing (Haviland & Clark, 1974; Prince, 1981). Given information includes words, concepts, and ideas that have already been mentioned in the discourse; new information involves words, concepts and ideas that have not yet been mentioned, and builds on the given information or launches a new thread of ideas. In the current research, the extent to which learners provide new information, compared to referring to previously shared information, will be captured by a measure called ***newness***.

In addition to information sharing, the team performance literature also advocates for concise communication between group members (Gorman, Cooke, & Kiekel, 2004; Gorman, Foltz, Kiekel, Martin, & Cooke, 2003). This is one of the reasons that formal teams, like military units, typically adopt conventionalized terminology and standardized patterns of communication (Salas et al., 2007). It is suggested that this concise communication is possible when there is more common ground within the team and the presence of shared mental models of the task and team interaction (Klein, Feltovich, Bradshaw, & Woods, 2005). The ***communication density*** measure used in the current research, was first introduced by Gorman et al. (2003) in team communication analysis to measure the extent to which a team conveys information in a concise manner. Specifically, the rate of meaningful discourse is defined by the ratio of semantic content to number of words used to convey that content. Using this measure, we will be able to further characterize the social roles that participants take on during collaborative interactions.

Taken together, we see that the sociocognitive processes involved in collaboration are internal to the individual but they are also manifested in the interactions with others in the group (Stahl, 2010). In particular, during group interactions, participants need to self-regulate their own learning and contributions, and co-regulate the learning and contributions of their collaborative partners. Reciprocally, the discourse of group members influences each participants' own monitoring and cognition (Chan, 2012; Järvelä, Hurme, & Järvelä, 2011). The social roles explored in this research are not necessarily reducible to processes of individual minds nor do they imply the existence of some sort of group mind. Rather, they are characterized by and emerge from the sequential interaction and weaving of semantic relations within a group discourse. The artifacts of transcribed communication resulting from collaborative interactions provides a window into the cognitive and social processes that define the participants' social roles. Thus, communication among the group members can be assessed to provide measures of participation, social impact, internal cohesion, responsiveness, newness, and communication density. These measures, which make up the GCA framework, define a space that can encompasses many key attributes of a collaborative group interaction. As participants exhibit more or less internal cohesion, responsiveness, social impact, new information, and communication density, we can associate them with a unique point in that space. As we will show, these points tend to cluster into distinct regions, corresponding to distinct patterns in behavioral engagement style and contribution characteristics. As such, these clusters represent characteristic roles that individuals take on during collaborative interactions, and they have an substantial impact on the overall success of those interactions.

## Construction of Group Communication Analysis (GCA) and Group Performance Measure

Transcripts of online group conversations provide two principle types of data (Foltz & Martin, 2009). First, transcript meta-data describe the patterns of interactions among group members. This includes who the participants are, the number, timing and volume of each of their contributions. Second is the actual textual content of each individual contribution, from which we can calculate the semantic relationships amongst the contributions. This involves taking semi-unstructured log file data, as depicted in Figure 1, and performing various transformations in order to infer the semantic relationship among the individual contributions, as depicted in Figure 2.

| | A | B | C | D |
|---|---|---|---|---|
| 1 | Group_id | person_id | chat_time | chat_text |
| 2 | 958 | 941 | 11/9/15 17:12 | hello |
| 3 | 958 | 347 | 11/9/15 17:13 | ok cool, everyone's here. sooo first question |
| 4 | 958 | 347 | 11/9/15 17:13 | ok so the certain characteristics to be considered to have a personality |
| 5 | 958 | 152 | 11/9/15 17:14 | okay so certain characterstics: doesn't it have to be like a stable thing? |
| 6 | 958 | 941 | 11/9/15 17:14 | I think the main thing about having a disorder is that its disruptive socially and/or makes the person a danger to himself or others |
| 7 | 958 | 343 | 11/9/15 17:14 | yes, stable over time |
| 8 | 958 | 152 | 11/9/15 17:14 | yeah, and it also mentioned it can't be because of drugs |
| 9 | 958 | 347 | 11/9/15 17:15 | also they have to have like unrealistic fantasies |
| 10 | 958 | 514 | 11/9/15 17:15 | yeah and not normal in their culture |
| 11 | 958 | 941 | 11/9/15 17:15 | no drugs or physical injury |
| 12 | 958 | 343 | 11/9/15 17:15 | begins in early adulthood or adolescence |
| 13 | 958 | 152 | 11/9/15 17:15 | i think that covers them? haha |
| 14 | 958 | 347 | 11/9/15 17:15 | ok, so arrogance doesn't just define it, they have to have most of these |
| 15 | 958 | 347 | 11/9/15 17:15 | yeah i think we got them |
| 16 | 958 | 152 | 11/9/15 17:15 | is it most or is it like 6? |
| 17 | 958 | 152 | 11/9/15 17:16 | because that was a question I wasn't sure of haha |
| 18 | 958 | 941 | 11/9/15 17:16 | there asking about disorders in general not just narcisism |
| 19 | 958 | 152 | 11/9/15 17:16 | Okay so second question? |
| 20 | 958 | 941 | 11/9/15 17:17 | the reading mentioned that some people may actually be as good as |
| 21 | 958 | 152 | 11/9/15 17:17 | yeahh it did |
| 22 | 958 | 152 | 11/9/11 17:17 | also, the person can't JUST be grandiose, they also need like praise and |
| 23 | 958 | 152 | 11/9/11 17:18 | others* |

Figure 1. Depiction of semi-unstructured log file data that is a typical artifact of CSCL interactions.

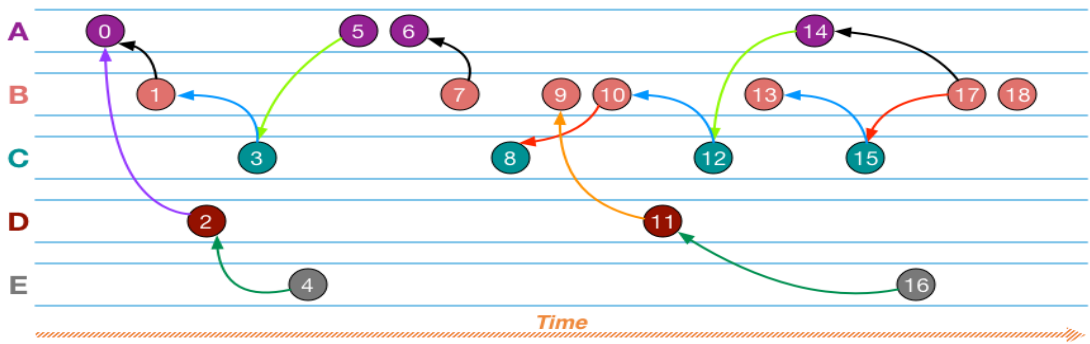

Figure 2. Schematic representation of inferring the semantic relationship among students' contributions in group interactions. The letters (i.e., A, B, C, D, E) on the vertical axis refer to students within a group interaction, and the numbers represent the sequential order of their discourse contributions.

Conversations, including collaborative discussions, commonly follow a statement-response structure, in which new statements are made in response to previous statements, and subsequently trigger further statements in response (see Figure 2). The structure of different online communications and discussion systems provide different affordances to the analyst to attribute a specific contribution as a response to some prior contribution. Regardless of the structure of the system, participants may, in a single contribution, refer to concepts and content presented in multiple previous contributions, made throughout the conversation either by themselves or other group members. Thus, a single contribution may be in response, to varying degrees, to many previous contributions, and it may in turn trigger, to varying degrees, multiple subsequent responses.

The analytical approach of the GCA was inspired by analogy to the cross- and auto-correlation measures from time-series analysis. Standard correlation measures how likely two variables are to be related. Cross-correlation similarly measures the relatedness between two variables, but with a given interval of time (or lag) between them. That is, for variables $x$ and $y$, and a lag of $\tau$, the cross-correlation would be the correlation of $x(t)$ with $y(t + \tau)$, across all applicable times, $t$, in the time-series. Standard correlation can be seen as a special case of the cross-correlation where $\tau = 0$, and the auto-correlation is a special case of one variable correlated with itself shifted in time ($\tau \neq 0$). By plotting the values of a cross correlation at different values of $\tau$ (typically from 1 up to some reasonably large value), one can identify if there are any statistically significant time-dependent relationship between the variables being examined. Such cross-correlation plots are a commonly used in the qualitative exploration of time series data.

While we might apply standard auto- and cross-correlation to examine temporal patterns in *when* participants contribute, we are primarily interested in understanding the temporal dynamics of *what* they contribute, and what the evolution of the conversation's semantics can teach us about the group's collaboration. To this end, a fine-grained measure of the similarity of participants' contributions is needed to capture the multi-responsive and social impact dynamics that may be present in collaborative interactions. There are different techniques for calculating the semantic similarity between two contributions. Two popular methods are content word overlap (CWO) and Latent Semantic Analysis (LSA). Each have their own strengths and weaknesses (Hu, Cai, Wiemer-Hastings, Graesser, & McNamara, 2007), however, these methods typically produce comparable results. In this research, similarity is measured using LSA. The semantic cohesion of contributions at fixed lags in the conversations can be computed much in the same way that cross-correlation evaluates correlation between lagged variables. Various measures of this auto- and cross-cohesion form the basis of the GCA's responsivity measures.

In addition to the GCA measures, the identification of topics covered in the group discussion affords us an objective measure of the overall group performance that is independent of the individual student performance (i.e., pre- and post-test scores). In the sections below, we describe the technical details of the construction of each of the GCA measures and the group performance measure (i.e., Topic Relevance).

**Participation measures.** The chat logs of a group discussion can be thought of as a sequence of individual contributions (i.e., verbal expressions within a conversational turn). In this sense, the boundaries of a contribution are defined by the nature of the technology that mediates the group discussion. A single contribution is a single message transmitted from one user to a group of others by way of a messaging service, or a single posting by a single user to a discussion forum. There may be multiple speech acts within a single contribution, but these will be treated as a single contribution. Further, a single user may transmit further contributions, immediately subsequent to their first, but these will be treated as separate contributions. So, the primary unit of analysis is a single contribution from a single user.

Let $C$ represent the sequence of contributions, with $c_t$ representing the $t^{th}$ contribution in the sequence. Let $n = |C|$ denote the length of the sequence. Since contributions represent turns in the discussion over time, the variable $t$ will be used to index individual contributions and will also be referred to as "time". The values of $t$ will range from 1 to $n$:

$$t \in \mathbb{Z}; \ 1 \leq t \leq n \tag{1}$$

Let $P$ be the set of participants in the discussion, of size $k = |P|$. Variables $a$ and $b$ in the following will be used to refer to arbitrary members (participants) in this set. In order to identify the contributor (or participant) that originated each statement, we define the following participation function, as depicted in Equation 2:

$$p_a(t) = \begin{cases} 1, & \text{if contribution } c_t \text{ was made by participant } a \in P \\ 0, & \text{otherwise} \end{cases} \tag{2}$$

The participation function for any participant, $a$, effectively defines a sequence, $P_a$:

$$P_a = \{p_a(t)\}_{t=1}^n = \{p_a(1), \ p_a(2), p_a(3), \dots, p_a(n)\} \tag{9}$$

of the same length, $n$, as the sequence of contributions $C$, which has the value 1 whenever participant $a$ originated the corresponding contribution in $C$, and 0 everywhere else. Using this participation function, it is relatively simple to define several useful descriptive measures of participation in the discussion. The number of contributions made by any participant is:

$$\|P_a\| = \sum_{t=1}^{n} p_a(t) \tag{3}$$

The sample mean participation of any participant is the relative proportion of their contributions out of the total:

$$\bar{p}_a = \frac{1}{n}\|P_a\| \tag{4}$$

and the sample variance in that participation is:

$$\sigma_a^2 = \frac{1}{n-1}\sum_{t=1}^{n}(p_a(t) - \bar{p}_a)^2 \tag{5}$$

If every participant contributed equally often, say by taking turns in round-robin fashion, then, for every participant:

$$\|P_a\| = \frac{n}{k} \tag{6}$$

This, in turn would result in mean participation scores of:

$$\bar{p}_a = \frac{1}{n}\cdot\frac{n}{k} = \frac{1}{k} \tag{7}$$

which will naturally get smaller for larger groups (larger $k$). This means that mean participation, $\bar{p}_a$, is not a useful measure of individual participation if we are comparing across groups of differing sizes. Therefore, we use the following measure to characterize individual participation:

$$\hat{p}_a = \bar{p}_a - \frac{1}{k} \tag{8}$$

This gives the mean participation above or below what we might expect from a perfectly equal participation. In the case where every participant contributes equally (equation 7), this measure is 0. A participant that contributes more than this will have a positive score, one that contributes less, a negative score. We refer to this as the group-relative mean participation.

We can, equivalently, represent the sequences of all participant as a $k \times n$ matrix, $\boldsymbol{M}$, by stacking the $k$ participation sequences as rows, in any arbitrary ordering (such that $i$ is an index over participants). Under this representation, the $(i,j)^{th}$ entry of the matrix is the $j^{th}$ contribution of participant $i$:

$$\boldsymbol{M}_{ij} = p_i(j); \ i \in P, 1 \leq j \leq n \tag{10}$$

By the definition of contributions given above, each contribution $c_t$ was originated by one and only one participant, so the participation function, $p_t$, will take on a value of 1 for exactly 1 participant at each time $t$, and be 0 for all other participants. One can see that the product of participation for different participants at the same time must always be 0:

$$p_a(t) \cdot p_b(t) = 0; \ a \neq b \tag{11}$$

It follows from (11) that the sum of each column in the matrix (10) would be exactly 1.

Since each participation sequence is, in effect, a time series of participant contributions, our goal to characterize the interactions between participants is a problem of characterizing their corresponding participation time series. The field of time series analysis gives us tools that we can either use directly or adapt to our needs. Specifically, we can make use of the cross-correlation between any two participants $a$ and $b$:

$$\rho_{a,b}(\tau) = \frac{1}{(n-1)\sigma_a \cdot \sigma_b} \sum_{t=\tau+1}^{n} p_a(t) \cdot p_b(t-\tau) - n \cdot \bar{p}_a \cdot \bar{p}_b \tag{12}$$

where the variable $\tau$:

$$\tau \in \mathbb{Z}; \ \tau \geq 0 \tag{13}$$

is some interval of time (or "lag") between the initial contribution of $b$ and then some subsequent contribution of $a$. A lag-1 cross-correlation between two participants will give a measure of how frequently one participant contributes immediately after the other participant. A lag-2 cross-correlation will give a measure of the responsiveness of the one participant after a single intervening contribution. One can qualitatively examine temporal patterns in any pair of participants' contributions by plotting this function for some reasonable number of lags. By looking at these plots for all pairs of users, one can examine the patterns for the entire group.

**Latent semantic analysis.** LSA represents the semantic and conceptual meanings of individual words, utterances, texts, and larger stretches of discourse based on the statistical regularities between words in a large corpus of natural language (Landauer, McNamara, Dennis, & Kintsch, 2007). The first step in LSA is to create a word-by-document co-occurrence matrix, in which each row represents a unique word and each column represents a "document" (in practice this typically means a sentence, paragraph, or section of an actual document). The values of the matrix represent counts of how many occurrences there were of each word in each document. For example, if the word "dog" appears once each in documents 1 and 9 and twice in document 50, and is considered the first word in the dataset, then the value of 1 will be in cells (1,1) and (1,9), and the value of 2 in cell (1,50). The occurrence matrix will then be weighted. Each row is weighted by a value indicating the importance of the corresponding word. Functional words (or "stop words") that occur with nearly even frequency across all documents receive small weights, since they are less useful at distinguishing documents. By contrast, words that have very different occurrences across the documents, and hence indicate more meaningful content terms, get higher weights. The most widely used weighting methods are term-frequency inverse document-frequency (TF-IDF) and Log-Entropy. A principal components analysis (PCA) is then performed on the weighted matrix by means of singular-value decomposition (SVD) matrix factorization. PCA is a procedure that allows one to reduce the dimensionality of a set of data such that it minimizes distortions in the relationships of the data. In the context of LSA, PCA allows us to reduce the word-by-document matrix to approximately 100-500 functional dimensions, which represent in compact form the most meaningful semantic relationships between words. The SVD procedure also yields a matrix which can be used to map the words from the original text corpus into vectors in a semantic space described by these semantic dimensions (i.e., LSA space).

When building an appropriate LSA space, it is necessary to have a corpus that broadly covers the topics under investigation. The Touchstone Applied Science Associates (TASA) corpus is a good example of a comprehensive set of tens of thousands of texts across numerous subject areas and spanning a range of levels of complexity (grade levels), which is suitable for building a general semantic space. In some instances, however, researchers desire a more custom corpus covering a specific domain, which is the case in the current research. The source corpora used in this research are conversational transcripts of collaborative interactions, which are not large enough to construct an LSA space. Furthermore, these transcripts refer to ideas and concepts that are not explicitly described in the transcripts. To obtain an appropriate representation of the semantic space we need to include external material that covers the topics of the conversations. One way to handle this problem is to enrich the source corpus with additional material that can provide appropriate background knowledge for key terms represented in the conversational transcripts (Cai, Li, Hu, & Graesser, 2016; Hu, Zhang, Lu, Park, & Zhou, 2009). The process begins with collecting a "seed" corpus of representative material (Cai, Burkett, Morgan, & Shaffer, 2011). In the current research, this included the chat transcripts for each data set, and the associated assigned reading material for students. This was done separately for each of the three datasets (described in the Methods section) to produce a custom domain specific seed corpus. This seed corpus is then scanned for key terms, which are used to scan the internet for documents (i.e., Wikipedia articles) on the topics mentioned in the seed corpus. The identified documents are used to create the expanded LSA space that is more comprehensive than the underlying transcripts on their own. For details on the extended LSA spaces for each of the corpora used in this research, please see the supplementary material.

By translating text into numerical vectors, a researcher can then perform any number of mathematical operations to analyze and quantify the characteristics of the text. One key operation is to compute the semantic similarity between any two segments of text. In the context of interactive chat, the similarity contributions $c_t$ and $c_u$ (where $u$, like $t$, is an index over time), can be computed by first projecting them into the LSA space, yielding corresponding document vectors $\vec{d}_t$ and $\vec{d}_u$. The projection is done by matching each word or term that occurs in the contribution, and locating the normalized term-vector for that word (calculated by the SVD process). These vectors are added together to get a vector corresponding to the entire contribution. If any term does not occur in the LSA space, it is ignored, and so does not contribute to the resulting vector. However, the construction of the space is such that this is very rare. Then, the cosine similarity of textual coherence (Dong, 2005), is computed on the document vectors $\vec{d}_t$ and $\vec{d}_u$, as described in equation 14. The cosine similarity ranges from -1 to 1, with identical contributions having a similarity score of 1 and completely non-overlapping contributions (no shared meaning) having a score of 0 or below (although in practice, negative text similarity cosines are very rare).

$$\cos(\vec{d}_t, \vec{d}_u) = \frac{\vec{d}_t \cdot \vec{d}_u}{\|\vec{d}_t\| \cdot \|\vec{d}_u\|}$$

(14)

The primary assumption of LSA is that there is some underlying or "latent" structure in the pattern of word usage across contexts (e.g., turns, paragraphs or sentences within texts), and that the SVD of the word-by-document frequencies will approximate this latent structure. The method produces a high-dimensional semantic space into which we can project participant contributions and measure the semantic similarity between them.

Using this LSA representation, students' contributions during collaborative interactions may be compared against each other in order to determine their semantic relatedness, and additionally, assessed for magnitude or salience within the high-dimensional space (Gorman et al., 2003). When used to model discourse cohesion, LSA tracks the overlap and transitions of meaning of text segments throughout the discourse.

Using this semantic relatedness approach, the semantic similarity score of any pair of contributions can be calculated as the cosine of the LSA document-vectors corresponding to each contribution. This works well as a measure of similarity between pairs of contributions. However, it must be aligned with the participation function in order to get a measure of the relationship between those participants in the discussion. As has been demonstrated above, the participation function can be used to select pairs of contributions related to a specific participant-participant interaction, and will screen out all other pairs of interactions. We therefore define a semantic similarity function:

$$s_{ab}(t,u) = p_a(t) \cdot p_b(u) \cdot \cos(\vec{d}_t, \vec{d}_u) \tag{15}$$

This will be the semantic similarity for contributions $c_t$ and $c_u$ only when contribution $c_t$ was made by participant $a$, and $c_u$ was made by participant $b$; otherwise it is zero (because in this case either $p_a(t)$ or $p_b(u)$, or both, would be 0). This product will form the foundation of several novel measures to characterize different aspects of participant involvement in the group discussion: the general participation, responsivity, internal cohesion, and social impact. These measures, described below, will be aligned and compared with Strijbos and De Laat (2010) conceptual framework to identify participant roles.

**Cross-cohesion.** This measure is similar in construction to the cross-correlation function, in that it assesses the relatedness of two temporal series of data to each other, at a given lag, $\tau$, though it relies on semantic cohesion rather than correlation as the fundamental measure of relatedness. This measure captures how responsive one participant's contributions are to another's over the course the collaborative interactions. Cross-cohesion is defined by averaging the semantic similarity of the contributions of the one participant to those of the other when they are lagged by some fixed amount, $\tau$, across all contributions:

$$\xi_{ab}(\tau) = \begin{cases} 0, & \|p_{ab}(\tau)\| = 0 \\ \dfrac{1}{\|p_{ab}(\tau)\|} \displaystyle\sum_{t=\tau+1}^{n} s_{ab}(t-\tau, t), & \|p_{ab}(\tau)\| \neq 0 \end{cases} \tag{16}$$

It is normalized by the total number of $\tau$-lagged contributions between the two participants, as expressed in equation 17:

$$\|p_{ab}(\tau)\| = \sum_{t=\tau+1}^{n} p_a(t-\tau) \cdot p_b(t) \tag{17}$$

We use the Greek letter $\xi$ (xi) to signify the cross-cohesion function. We refer to $\xi_{ab}(\tau)$ as the "cross-cohesion of $a$ to $b$ at $\tau$" or as the "$\tau$-lagged cross-cohesion of $a$ to $b$". The cross-cohesion function measures the average semantic similarity of all $\tau$-lagged contributions between two participants. As such, it gives an insight into both the degree and rapidity to which one participant may be responding to the comments of another. The first participant (the initiator), denoted by $a$, is that user whose prior contribution potentially triggered a subsequent response. The second participant (the respondent), denoted by $b$, is that user whose contribution potentially responds to some part of the initiator's contribution. The cross-cohesion at 1 represents the propensity for respondents to respond to the content of the initiator's immediately previous contribution. The propensity for respondents to comment on an initiator's contribution after 1 intervening contribution is characterized by the 2-lagged cross-cohesion matrix, and so on. In the special case that the initiator and the respondent are the same, we may refer to this as the auto-cohesion function, in similar fashion to the auto-correlation function. The auto-cohesion function measures consistency over time in the semantics of a single participant's contributions. The most similar work to date (Samsonovich, 2014) has made use of the standard cross-correlation function applied to timeseries of numeric measures computed from natural language, and then draws inferences from these as to the nature of social interactions. However, the use of semantic similarity as the base measure of relatedness in lieu of correlation, is, to our knowledge, entirely novel.

**Responsivity.** Cross-cohesion at a single lag may not be very insightful on its own, as it represents a very narrow slice of interaction. By averaging over a wider window of contributions, we can get a broader sense of the interaction dynamics between the participants.

For a conversation with $k = |P|$ participants, and given some arbitrary ordering of participants in $P$, we can represent cross-cohesion as a $k \times k$ matrix $\boldsymbol{X}(\boldsymbol{\tau})$, such that the element in row $i$, column $j$ is given by the cross-cohesion function $\xi_{ij}(\tau)$. We refer to this matrix as "$\tau$-lagged cross-cohesion matrix", or "cross-cohesion at $\tau$". The rows of the matrix represent the responding students, who we refer to as the respondents. The columns of the matrix represent the initiating students, referred to as the initiators. We define responsivity across a time window as follows:

$$\boldsymbol{R}(\boldsymbol{w}) = \frac{1}{w} \sum_{\tau=1}^{w} \boldsymbol{X}(\boldsymbol{\tau}) \tag{18}$$

This will be referred to this as "w-spanning responsivity" or "responsivity across w". An individual entry in the matrix, $r_{ab}(w)$ is the "w-spanning responsivity of a to b" or the "responsivity of a to b across w". These measures form a moving-average of responsivity across the entire dialogue. The window for the average consists of a trailing subset of contributions, starting with the most current and looking backwards over a maximum of w prior contributions. Characteristics of an individual participant can be obtained by averaging over their corresponding rows or columns of the w-spanning responsivity matrix, and by taking their corresponding entry in the diagonal of the matrix. For details on the spanning window calibration used for the datasets in the current research, please see the supplementary material.

***Internal cohesion.*** Internal cohesion is the measure of how semantically similar a participant's contributions are with their own previous contributions during the interaction. The participant's "w-spanning internal cohesion" is characterized by the corresponding diagonal entry in the w-spanning responsivity matrix:

$$r_{aa}(w) \tag{19}$$

The internal cohesion is effectively the average of the auto-cohesion function of the specified participant over the first *w lags*.

***Overall responsivity.*** Each row in the w-spanning responsivity matrix is a vector representing how the corresponding participant has responded to all others. To characterize how responsive a participant is to all other group members' contributions during the collaborative interactions, we take the mean of these row vectors (excluding the participant of interest):

$$\bar{r}_a(w) = \frac{1}{k-1} \sum_{i=1; i\neq a}^{k} r_{ai}(w) \tag{20}$$

This is referred to as the "w-spanning responsivity of a", or just the "overall responsivity of a" for short.

**Social impact.** Each column in the w-spanning responsivity matrix is a vector representing how contributions initiated by the corresponding participant have triggered follow-up responses. In a similar fashion to the overall responsivity described above, a measure of each individual participant's social impact can be calculated by averaging over these column-vectors (excluding the participant of interest):

$$\bar{\iota}_a(w) = \frac{1}{k-1} \sum_{j=1; j\neq a}^{k} r_{ja}(w) \tag{21}$$

This is referred to as the "w-spanning impact of a", or just the "social impact of a" for short.

**LSA Given-New.** Participants' contributions can vary in how much *new* versus *given* information they contain (Hempelman et al., 2005; McCarthy et al., 2012). Note that, for the purposes of the current research, we were more interested in a measure of the amount of new rather than given information provided by participants. This is motivated by the fact the responsivity measures already capture the social equivalent of "givenness", which is more relevant in the contexts of group interactions. Establishing how much new information is provided in any given contribution can be meaningful to the dynamics of the conversation, as well as to characterize the ways in which different participants contribute. Following the method of Hu et al., 2003, the given information at the time of contribution $t$ is a subspace of the LSA spanned by the document vectors of all previous contributions:

$$G_t = span\{\vec{d}_1, \vec{d}_2, \dots, \vec{d}_{t-1}\} \tag{22}$$

The semantic content of the current contribution can then be divided into the portion already given by projecting the LSA document vector for the current contribution onto the subspace defined in equation (23):

$$\vec{g}_t = Proj_{G_t}(\vec{d}_t) \tag{23}$$

There is also the portion of semantic content that is new to the discourse by projecting the same document vector onto the orthogonal complement of the given subspace, as defined in (24):

$$\vec{n}_t = Proj_{G_t^{\perp}}(\vec{d}_t) \tag{24}$$

This is the portion perpendicular to the given subspace. Of course, the semantic content of the contribution is completely partitioned by these projections, so:

$$\vec{d}_t = \vec{g}_t + \vec{n}_t \tag{25}$$

In order to get a useful measure of the total amount of new semantic content provided in any given contribution, we take the relative proportion of the size of the given vector to the total content provided:

$$n(c_t) = \frac{\|\vec{n}_t\|}{\|\vec{n}_t\| + \|\vec{g}_t\|} \tag{26}$$

This given-new value ranges between 0 (all given content, nothing new) to 1 (all new content).

*Newness.* We can characterize the relative new content provided by each individual participant by averaging over the given-new score of their contributions:

$$\overline{N}_a = \frac{1}{\|P_a\|} \sum_{t=1}^{n} p_a(t) \cdot n(c_t) \tag{27}$$

**Communication Density.** Another meaningful measure involves calculating the average amount of semantically meaningful information provided in a contribution. This measure was first established by Gorman et al. (2003) in their work examining team communication in a synthetic military aviation task. This measure differs from the **Given-New** measure in that it is entirely calculated from the contribution $c_i$ and its corresponding LSA vector, $\vec{d}_i$, and does not consider any prior contributions. The communication density is defined in (28):

$$D_i = \frac{\|d_i\|}{\|c_i\|} \tag{28}$$

where $\|d_i\|$ is the norm of the LSA vector and $\|c_i\|$ is the length of the contribution in words. Thus, communication density gives the per-word amount of semantic meaning for any contribution. In order to characterize the communication density of a particular participant, we must calculate the average density over all of their contributions:

$$\overline{D}_a = \frac{1}{\|P_a\|} \sum_{t=1}^{n} p_a(t) \cdot D_t \qquad (29)$$

The six measures that comprise the GCA are summarized in Table 2:

**Table 2** Collaborative Interaction Process Measures from the GCA

| Measure | Equation | Description |
|---|---|---|
| Participation | (8) | Mean participation of any participant above or below what you would expect from equal participation in a group of the size of theirs |
| Overall Responsivity | (20) | Measure of how responsive a participant's contributions are to all other group members' recent contributions |
| Internal cohesion | (19) | How semantically similar a participant's contributions are with their own recent contributions |
| Social impact | (21) | Measure of how contributions initiated by the corresponding participant have triggered follow-up responses |
| Newness | (27) | The amount of new information a participant provides, on average |
| Communication density | (29) | The amount of semantically meaningful information |

## Topic Modeling

The cohesion-based discourse measures described above capture important intrapersonal and interpersonal dynamics, but an additional data mining technique is needed to capture the themes and topics of the collaborative discussions. The identification of covered topics is of particular interest for the current analyses because it affords an assessment of the overall group performance that is independent of the individual student performance (i.e., pretest and post-test scores). Latent Dirichlet allocation (LDA; Blei, Ng, & Jordan, 2003), more commonly known as "topic modeling" (Steyvers & Griffiths, 2007), is a method of deriving an underlying set of topics from an unlabeled corpus of text.

Topic modeling allows researchers to discover the common themes in a large body of text, and to what extent those themes are present in individual documents. Topic modeling has frequently been used to explore collaborative learning contexts (e.g., Cai et al., 2017). In this research, LDA topic models were used to provide an inference mechanism of underlying topic structures through a generative probabilistic process. This generative process delivers a distribution over topics for each document in the form of a proportion. This distribution can be used to find the topics most representative of the contents of that document. These distributions can also be considered as data for future analyses, as every document's distribution describes a document-topic "fingerprint". For this research, the topic model corpus for each of the three data sets (described in the Methods section) consisted of the extended corpora produced with the "seed method" described earlier (see the LSA section above). A topics model was then generated for each of these extended corpora. The identified topics were inspected to identify which topics might be considered "off-task" for the corresponding collaborative activity (details of this are described in the Methods section). Thus, the topics were divided into two groups, namely domain content relevant and irrelevant.

**Topic Relevance**. The measure of group performance was operationalized as the amount of on-topic discussion. To develop a meaningful measure of relevant or "on-task" discussion, we begin with the set of all topics, $Q$, constructed as described above. The topic score:

$$t_q(c_t) \tag{30}$$

gives the proportion of contribution $c_t$ that covers topic $q \in Q$. These proportions sum to 1 for any contribution:

$$\sum_{q \in Q} t_q(c_t) = 1 \tag{31}$$

The set of all topics will be manually partitioned into two subsets, $Q'$ and $Q°$:

$$Q = Q' \cup Q°; \; Q' \cap Q° = \{\emptyset\} \tag{32}$$

$Q'$ represents those topics considered "relevant" or "on-task" for the corresponding collaborative activity, and $Q°$ consists of all other "off-task" topics (see Methods section). We can then construct a measure of the relative proportion of on-task material in each contribution by summing over the topic scores for topics in $Q'$:

$$T'(c_t) = \sum_{q \in Q'} t_q(c_t) \tag{33}$$

We can get a measure of the degree to which the entire group discussion was on or off-task by averaging this across the entire discussion:

$$T' = \frac{1}{n} \sum_{t=1}^{n} T'(c_t) \tag{34}$$

# Methods

The GCA measures (as summarized in Table 2) were computed for each of three independent collaborative interaction datasets. The first, is a Traditional Computer-Supported Collaborative Learning (CSCL) dataset. It is important to note that the Traditional CSCL dataset will be the primary dataset used in the analyses. The second is a synchronous massive online course (SMOC) dataset called UT2014 SMOC. The third is a collaborative learning and problem-solving data set collected from a virtual internship game called "Land Science". The three datasets are described below.

## Traditional CSCL Dataset

**Participants.** Participants were enrolled in an introductory-level psychology course taught in the Fall semester of 2011 at a Southwest University. While 854 students participated in this course, some minor data loss occurred after removing outliers and those who failed to complete the outcome measures. The final sample consisted of 840 students. Females made up 64.3% of this final sample. 50.5% of the sample identified as Caucasian, 22.2% as Hispanic/Latino, 15.4% as Asian American, 4.4% as African American, and less than 1% identified as either Native American or Pacific Islander.

**Course Details and Procedure.** Students were told that they would be participating in an assignment that involved a collaborative discussion on personality disorders, as well as several quizzes. Students were told that their assignment was to log onto an online educational platform specific to the University at a specified time (Pennebaker, Gosling, & Ferrell, 2013). Students were also instructed that, prior to logging onto the educational platform, they should read certain assigned material on personality disorders.

After logging onto the system, students took a 10-item, multiple choice pretest quiz. This quiz asked students to apply their knowledge of personality disorders to various scenarios and to draw conclusions based on the nature of the disorders. After completing the quiz, they were randomly assigned to a chatroom with one to four classmates, also chosen at random (average group size was 4.59), and instructed to engage in a discussion of the assigned material. The group chat began as soon as someone typed the first message and lasted for exactly 20 minutes, when the chat window closed automatically. Then students took a second 10 multiple-choice question post-test quiz. Each student contributed 154.0 words on average ($SD = 104.9$) in 19.5 sentences ($SD = 12.5$). As a group, discussions were about 714.8 words long ($SD = 235.7$) and 90.6 sentences long ($SD = 33.5$).

**Group Performance Measure.** The group performance was operationally defined as the proportion of topic-relevant discussion during the collaborative interaction, as described in Equation 34. As a reminder, the corpus used for the topic modeling was the same extended corpus (created using the seed method described earlier) used for creating the custom LSA spaces (Cai et al., 2011).

The topic modeling analysis revealed twenty topics, of which eight were determined to be relevant to the collaborative interaction task. Interjudge reliability was not used to determine the relevant topics. Instead, two approaches were used to determine the most relevant topics and validate a topic relevance measure for group performance. The first was the frequency of the topics discussed across all the groups and individual students, wherein more frequently discussed topics were viewed as more important. Second, correlations between the topics and student learning gains were used to help validate the importance of the topic. Once the important topics were determined, an aggregate topic relevance score was computed by summing up the proportions for those topics (Equation 33). The top 10 words for each of the relevant topics are listed in the supplementary material.

## UT2014 SMOC Dataset

**Participants.** Participants were 1,713 students enrolled in an online introductory-level psychology course taught in the Fall semester of 2014 at a Southwest University. Throughout the course, students participated in a total of nine different computer-mediated collaborative interactions on various introductory psychology topics. This resulted in a total of 3,380 groups, with four to five students per group. However, 83 out of 3,380 (2.46%) chat groups were dropped because they contained only a single participant.

**Course Details and Procedure.** The collaborative interactions took place in a large online introductory-level psychology course. The structure of the class followed a synchronous massive online course (SMOC) format. SMOCs are a variation of massive open online courses (MOOCs) (Chauhan, 2015). MOOCs are open to the general public and typically free of charge. SMOCs are limited to a total of 10,000 students, including those enrolled at the university and across the world, and are available to all the participants at a registration fee of $550 (Chauhan, 2015).

The course involved live-streamed lectures that required students to log in at specific times. Once students were logged onto the university's online educational platform, students were able to watch live lectures and instructional videos, take quizzes and exercises, and participate in collaborative discussion exercises. Students interacted in collaborative discussions via web chat with randomly classmates. Once put into groups, students were moved into a chat room and told they had exactly 10 minutes to discuss the assigned material (readings or videos). This 10 minute session began at the moment of the first chat message. At the end of the discussion, students individually took a 10-item, multiple choice quiz that asked students to apply their knowledge of the assigned material to various scenarios and to draw conclusions.

## Land Science Dataset

**Participants.** A total of 38 participants interacted in 19 collaborative problem-solving simulation games. Each game consisted of multiple rooms, and each room involved multiple chat sessions. There was a total of 630 distinct chat sessions. Of the 38 participants, $n = 29$ were student players, $n = 13$ were Mentors, $n = 10$ were Teachers, and $n = 1$ was a Non-Player Character (NPC). For the purposes of detecting the social roles of players, only the Players' and Mentors chat' were analyzed with the GCA. One of the rationales for exploring this dataset was to evaluate the generalizability of the GCA method across a range of different types of collaborative tasks. Specifically, unlike the collaborative learning datasets described above, Land Science is a collaborative problem-solving environment.

**Details and Procedure.** Land Science is an interactive virtual urban-planning internship simulation with collaborative problem solving (Bagley & Shaffer, 2015; Shaffer, 2006; Shaffer & Graesser, 2010). The goal of the game is for students to think and act like STEM professionals. Players are assigned an in-game role as an intern with a land planning firm in a virtual city, under the guidance of a mentor. During the game, players communicate with other members of their planning team, as well as with a mentor who sometimes role plays as a professional planning consultant. Players are deliberately given different instruction and resources; they must successfully combine skills within small teams in order to solve the collaborative problems.

## Detecting Social Roles

The following analyses focus on addressing the main questions raised in the Overview of Present Research, above. The analysis started with the Traditional CSCL dataset, which was immediately partitioned into training (84%) and testing (16%) datasets. Descriptive statistics for the GCA measures from the training data are presented in Table 3.

**Table 3** Descriptive Statistics for GCA Measures

| Measure | Minimum | Median | *M* | *SD* | Maximum |
|---|---|---|---|---|---|
| Participation | -0.26 | -0.01 | 0.00 | 0.10 | 0.35 |
| Social Impact | 0.00 | 0.18 | 0.18 | 0.05 | 0.43 |
| Overall Responsivity | 0.00 | 0.18 | 0.18 | 0.05 | 0.50 |
| Internal Cohesion | -0.06 | 0.18 | 0.18 | 0.09 | 0.58 |
| Newness | 0.00 | 0.48 | 0.78 | 1.25 | 18.09 |
| Communication Density | 0.00 | 0.21 | 0.34 | 0.51 | 6.45 |

*Note*. Mean (***M***). Standard deviation (***SD***)

The data were normalized and centered to prepare them for analysis. Specifically, the normalization procedure involved Winsorising the data based on each variable's upper and lower percentile. Density and pairwise scatter plots for the GCA variables are reported in the supplementary material. A cluster analysis approach was adopted to discover communication patterns associated with specific learner roles during collaborative interactions. Cluster analysis is a common data mining technique that involves identifying subgroups of data within the larger population who share similar patterns across a set of variables (Baker, 2010). Cluster analysis has been applied in previous studies of social roles (e.g., Lehmann-Willenbrock et al., 2016;

Risser & Bottoms, 2014) and has proven useful in building an understanding of individuals' behaviors in many digital environments more broadly (Mirriahi, Liaqat, Dawson, & Gašević, 2016; Valle & Duffy, 2007; Wise et al., 2012). Prior to clustering, multicollinearity was assessed through inflation factor (VIF) statistics and collinearity was assessed using Pearson correlations. The VIF results support the view that multicollinearity was not an issue with VIF > 7 (Fox & Weisberg, 2010). There was evidence of moderate collinearity between two variables, newness and communication density. However, further evaluation showed that collinearity did not impact the clustering results. For more details on collinearity and cluster tendency assessments, please see the supplementary material.

In principle, any number of clusters can be derived from a dataset. So, the most important decision for any analyst when making use of cluster analysis is to determine the number of clusters that best characterizes the data. There are several methods suggested in the literature for determining the optimal number of clusters (Han, Pei, & Kamber, 2011). A primary intuition behinds these methods is that ideal clusterings involve compact, well-separated clusters, such that the total intra-cluster variation or total within-cluster sum of square (wss) is minimized (Kaufman & Rousseeuw, 2005). In the current research, we used the NbClust R package which provides 26 indices for determining the relevant number of clusters (Charrad, Ghazzali, Boiteau, & Niknafs, 2014). It is beyond the scope of this paper to specify each index, but they are described comprehensively in the original paper of Charrad et al. (2014). An important advantage of NbClust is that researchers can simultaneously compute multiple indices and determine the number of clusters using a majority rule, wherein the proposed cluster size that has the best score across the majority of the 26 indices is taken to be optimal. Figure 3 reveals that the optimal number of clusters, according to the majority rule, was 6 for a K-means clustering. Note, a two and four-cluster solution was also inspected, and compared. An in-depth coverage of those models and their evaluation may be found in the supplementary material.

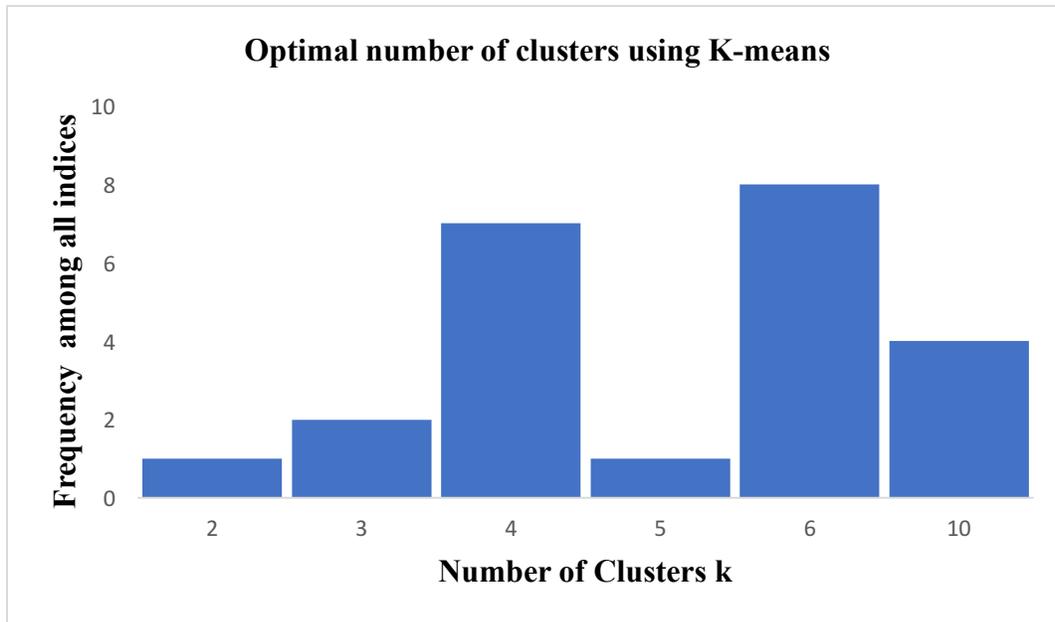

Figure 3. Frequency for recommended number of clusters using K-means, ranging from 2 to 10, using 26 criteria provided by the NbClust package. Here we see 8 of the 26 indices proposed 6 as the optimal number of clusters in the Traditional CSCL dataset.

## Cluster Analysis

K-means was used to group learners with similar GCA profiles into clusters. Investigation of the cluster centroids may shed light on whether the clusters are conceptually distinguishable. Centroids are representative of what may be considered typical, or average, of all entities in the cluster. With K-means, the centroids are in fact the means of the points in the cluster (although this is not necessarily true for other clustering methods). In the context of GCA profiles, we may interpret the centroids as behavior typical of a distinct style of interaction (i.e. roles). The centroids for the six-cluster k-means solution are presented in Figure 4, below. It is worth noting that since the clustering was performed on normalized data, zero in this figure represents the population average for each measure, while positive and negative values represent values above or below that average, respectively.

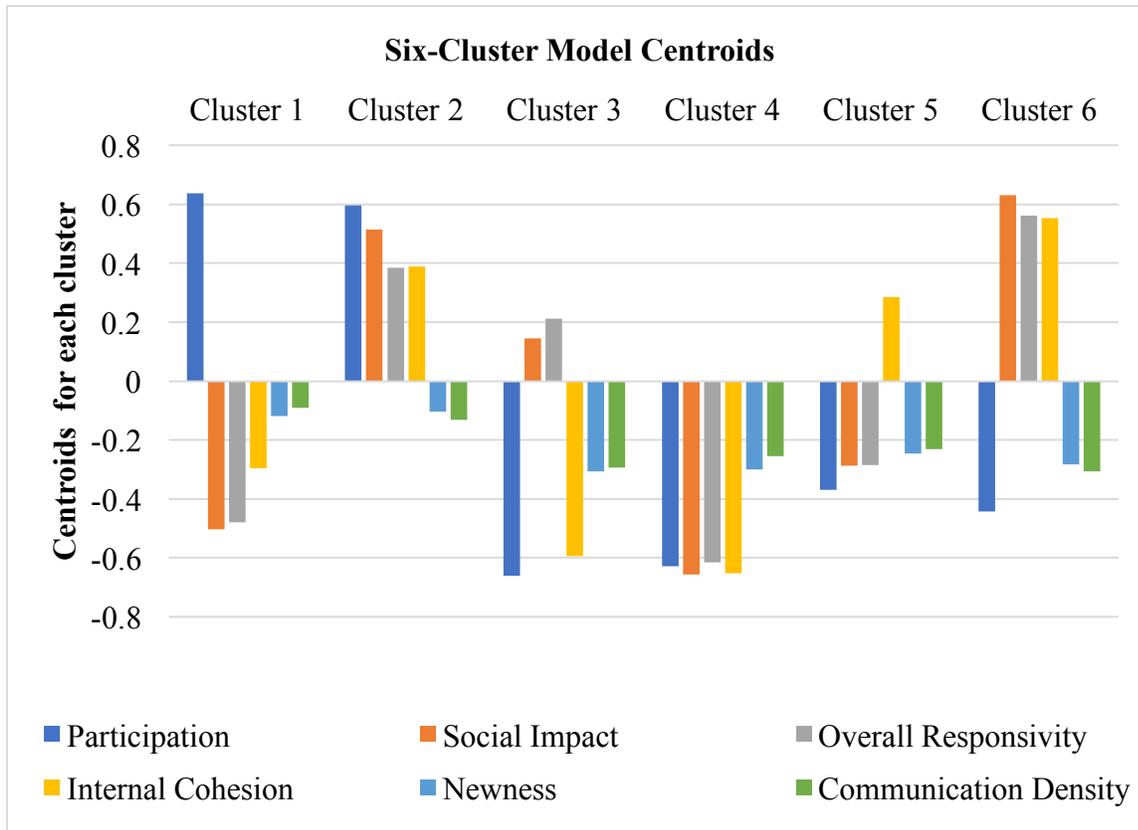

Figure 4. Centroids for the six-cluster solution across the GCA variables.

We see some interesting patterns across the six-cluster solution. Cluster 1 ($N$ = 143) was characterized by learners who have the highest participation, however very low scores across all other measures. This suggests that, when they contribute, their discourse is more in response to themselves than other team members since they exhibit relatively higher internal cohesion than responsiveness or social impact. Furthermore, their contributions do not seem to warrant further discussion from the group members or provide new information (i.e., low social impact and newness). These individuals are similar to the **Over-riders** described in Strijbos and De Laats' (2010) framework, and are labeled as such in the current research. Strijbos and De Laat (2010) described these individuals as exhibiting strong individual learning goals and attempting to push the group members into adopting their agenda. In contrast to the Drivers (cluster 2) role, Over-riders have a higher degree of internal cohesion compared to social impact or responsiveness, which signals the Over-rider is more concerned with a personal narrative than with collaboration or the prevailing social climate.

The learners in Cluster 2 ($N$ = 153) were among the highest participators; they exhibit high social impact, responsiveness, and internal cohesion, but coupled with the lowest newness and communication density. Learners in these clusters are investing a high degree of effort in the collaborative discussion and display self-regulatory and social-regulatory skills. This pattern is labeled the ***Driver*** in the current research.

Cluster 3 ($N = 88$) is characterized by learners who have the lowest participation. However, when they do contribute it appears to build, at least minimally, on previously contributed ideas and move the collaborative discourse forward (i.e., moderately positive social impact and responsiveness). This cluster is labeled as the ***Follower***.

Cluster 4 ($N = 117$), labeled the as the ***Lurker***, is characterized by some of the lowest values across all GCA measures. Lurkers have been defined differently in the literature, ranging from non-participators to minimal participators (Nonnecke & Preece, 2000; Preece, Nonnecke, & Andrews, 2004). The distinction between a Ghost and a Lurker is not clear, and the terms appear to be interchangeable, although Strijbos and De Laat (2010) do make a distinction based on group size. There are two reasons that motivated us to prefer the term Lurker, rather than Ghost, in the current research. First, the GCA methodology would not be able to detect an individual that did not participate at all (because there would not be a log file for those students), which suggests the learners in these clusters did contribute at least minimally. Second, past research has labeled the Ghost and Lurker roles predominantly based on the amount of contributions a student makes. However, the GCA captures participation as well as the sociocognitive characteristics of those contributions. Again, since these measures are normalized, the very low values for this cluster centroid does not suggest these students have no social impact, or were completely unresponsive to others. Rather it suggests that these students expressed far less, as compared to the population average. Lurking behavior sometimes involves some level of engagement but at other times little engagement so it is associated with both positive and negative outcomes in the literature (Preece et al., 2004). Therefore, Lurker appeared to be the most appropriate label for cluster 4.

Learners occupying cluster 5 ($N = 91$) exhibited high internal cohesion, but low scores on all the other GCA measures. This cluster is labeled as ***Socially Detached***, because the pattern appears to capture students who are not productively engaged with their collaborative peers, but instead focused solely on themselves and their own narrative.

In cluster 6 ($N = 126$), we see learners with low participation, but when they do contribute, they attend to other learners' contributions and provide meaningful information that furthers the discussion (i.e., high internal cohesion, overall responsiveness, and social impact). It is interesting to note that these students are not among the highest participators, but their discourse signals a social positioning that is conducive to a productive exchange within the collaborative interaction. This pattern is suggestive of a student that is engaged in the collaborative interaction, but takes a more thoughtful and deliberative stance, than do the **Drivers**. As such, we refer to this cluster as a ***Task-Leader*** in this research. Overall, the six-cluster model appears, at least upon an initial visual inspection, to produce theoretically meaningful participant roles. We then proceeded to evaluate the quality and validity of this model.

## Clustering Evaluation and Validation

The literature proposes several cluster validation indexes that quantify the quality of a clustering (Hennig, Meila, Murtagh, & Rocci, 2015). In principle, these measures provide a fair comparison of clustering and aid researchers in determining whether a particular clustering of the data is better than an alternative (Taniar, 2006). There are three main types of cluster validation measures and approaches available: internal, stability, and external. Internal criteria evaluate the extent to which the clustering "fits" the data set based on the actual data used for clustering. In the current research, two commonly reported internal validity measures (Silhouette and Dunn's index) were explored using the R package clValid (Brock, Pihur, Datta, & Datta, 2008). Silhouette analysis measures how well an observation is clustered and it estimates the average distance between clusters (Rousseeuw, 1987). Silhouette widths indicate how discriminating the candidate clusters are by providing values that range from a low of -1, indicating that observations are likely placed in the wrong cluster to 1, indicating that the clusters perfectly separate the data and no better alternative clustering can be found. The average silhouette (AS) for the six-cluster model was positive (AS = .31), indicating the students in a cluster had higher similarity to other students in their own cluster than to students in any other cluster. Dunn's index ($D$) evaluates the quality of clusters by computing a ratio between the inter-cluster distance (i.e., the separation between clusters) and the intra-cluster diameter (i.e., the within-cluster compactness). Larger values of D suggest good clusters, and a D larger than 1 indicates compact separated clusters (Dunn, 1974). Dunn's index for the six-cluster model was ($D$ = .5), indicating that this clustering has moderate compactness.

Stability is another important aspect of cluster validity. A clustering may be said to be stable if it's clusters should remain intact (i.e., not disappear easily) when the data set is changed in a non-essential way (Hennig, 2007). While there may be many different conceptions of what constitutes a "non-essential change" to a dataset, the leave-one-column out method is commonly applied. The stability measures calculated in this way compare the results from clustering based on the complete dataset to clusterings based on removing each column, one at a time (Brock et al., 2008; Datta & Datta, 2003). In the current context this corresponds to the removal of one of the GCA variables at a time. The stability measures are the average proportion of non-overlap (APN), the average distance (AD), the average distance between means (ADM), and the figure of merit (FOM). Each of these measures is calculated for each reduced dataset (produced by dropping one column), and their average is taken as the measure for the dataset as a whole. The APN ranges from 0 to 1, while the AD, ADM, FOM all range from 0 to infinity. For each of these measures, smaller scores indicate a better, more stable clustering. The stability scores for the six-cluster solution suggest the clusters are quite stable across the four measures with APN = .22, AD = .97, ADM = .31, and FOM= .37.

## Cluster Coherence

It is important to also evaluate the coherence of the clusters in terms of their underlying GCA variables. This can help to establish that the identified clusters do in fact represent distinct modes in the distribution of the GCA measures. Consequently, the six-cluster model was further evaluated to determine whether learners in the cluster groups significantly differed from each other on the six GCA variables. The multivariate skewness and kurtosis were investigated using the R package MVN (Korkmaz, Goksuluk, & Zararsiz, 2015) which produces the chi-square Q-Q plot (see supplementary material) and a Henze-Zirkler (*HZ*) test statistic which assesses whether the dataset follows an expected multivariate normal distribution. The results indicated the GCA variables did not follow a normal distribution, $HZ = 5.06$, $p < .05$. Therefore, a permutational MANOVA (or nonparametric MANOVA) was to evaluate between-cluster GCA variable means. The permutational MANOVA, implemented in the *Adonis* routine of the VEGAN package in R (Oksanen et al., 2016), is a robust alternative to both the traditional parametric MANOVA and to ordination methods for describing how variation is attributed to different experimental treatments or, in this case, cluster partitions (Anderson, 2001). The *Adonis* test showed a significant main effect of the clusters, $F(5,712)=350.86$, $p < .001$. These results support the models' formation and ability to organize learners based on differences in their collaborative communication profiles.

The analyses proceeded with ANOVAs, followed by Tukey's *post hoc* comparisons to identify significant differences in the participants' scores on the six GCA variables between the clusters. Levene's Test of Equality of Error Variances was violated for all the GCA variables so a more stringent alpha level ($p < .01$) was used when identifying significant differences for these variables (Tabachnick & Fidell, 2007, p. 86). The ANOVA main effect $F$-values along with the means and standard deviations for the GCA variables across each cluster are reported in Table 4 for the six-cluster model. The ANOVA revealed significant differences among clusters for all of the six GCA variables at the $p < .0001$ level for the six-cluster model. Tukey's HSD *post hoc* comparisons for the six-cluster model is Table 5, where we can see that the observed differences in GCA profiles across the clusters were, for the majority, significantly distinct in both models.

**Table 4** Six-cluster Model Means and Standard Deviations for the Six GCA Variables

| GCA Measures | Cluster 1: Over-rider $n$ = 143 $M(SD)$ | Cluster 2: Driver $n$ = 153 $M(SD)$ | Cluster 3: Follower $n$ = 88 $M(SD)$ | Cluster 4: Lurker $n$ = 117 $M(SD)$ | Cluster 5: Detached $n$ = 91 $M(SD)$ | Cluster 6: Task-Leader $n$ = 126 $M(SD)$ | $F$-value |
|---|---|---|---|---|---|---|---|
| Participation | 0.64 (0.23) | 0.60 (0.24) | -0.66 (0.28) | -0.63 (0.27) | -0.37 (0.36) | -0.44 (0.32) | 285.70*** |
| Social Impact | -0.50 (0.31) | 0.51 (0.33) | 0.15 (0.47) | -0.66 (0.23) | -0.29 (0.39) | 0.63 (0.25) | 200.50*** |
| Overall Responsivity | -0.48 (0.31) | 0.38 (0.38) | 0.21 (0.46) | -0.61 (0.24) | -0.28 (0.36) | 0.56 (0.28) | 157.70*** |
| Internal Cohesion | -0.30 (0.37) | 0.39 (0.31) | -0.59 (0.21) | -0.65 (0.17) | 0.29 (0.31) | 0.55 (0.23) | 210.30*** |
| Newness | -0.12 (0.14) | -0.10 (0.14) | -0.31 (0.14) | -0.30 (0.13) | -0.25 (0.15) | -0.28 (0.12) | 15.83*** |
| Communication Density | -0.09 (0.14) | -0.13 (0.16) | -0.29 (0.15) | -0.26 (0.15) | -0.23 (0.16) | -0.31 (0.12) | 15.01*** |

Note: ANOVA $df$ = 5,712; *** $p < .0001$

**Table 5** Tukey-HSD P-Values for the Pairwise Comparisons for the GCA Measures Across the Six-Cluster Solution

| Cluster Comparison | GCA Variables | | | | | |
|---|---|---|---|---|---|---|
| | Participation | Social Impact | Overall Responsivity | Internal Cohesion | Newness | Communication Density |
| 2 vs. 1 | $p = .04$ | $p < .001$ | $p < .001$ | $p < .001$ | $p = .83$ | $p = .06$ |
| 3 vs. 1 | $p < .001$ | $p < .001$ | $p < .001$ | $p < .001$ | $p < .001$ | $p < .001$ |
| 4 vs. 1 | $p < .001$ | $p < .001$ | $p < .001$ | $p < .001$ | $p < .001$ | $p < .001$ |
| 5 vs. 1 | $p < .001$ | $p = .008$ | $p = .05$ | $p < .001$ | $p < .001$ | $p < .001$ |
| 6 vs. 1 | $p < .001$ | $p < .001$ | $p < .001$ | $p < .001$ | $p < .001$ | $p < .001$ |
| 3 vs. 2 | $p < .001$ | $p < .001$ | $p = .66$ | $p < .001$ | $p < .001$ | $p < .01$ |
| 4 vs. 2 | $p < .001$ | $p < .001$ | $p < .001$ | $p < .001$ | $p < .001$ | $p < .01$ |
| 5 vs. 2 | $p < .001$ | $p < .001$ | $p < .001$ | $p = .58$ | $p < .05$ | $p < .001$ |
| 6 vs. 2 | $p < .001$ | $p = .07$ | $p < .001$ | $p < .001$ | $p < .001$ | $p < .001$ |
| 4 vs. 3 | $p = .93$ | $p < .001$ | $p < .001$ | $p = .99$ | $p = 1.00$ | $p = .99$ |
| 5 vs. 3 | $p < .001$ | $p < .001$ | $p < .001$ | $p < .001$ | $p = .56$ | $p = .50$ |
| 6 vs. 3 | $p < .001$ | $p < .001$ | $p < .001$ | $p < .001$ | $p = .99$ | $p = 1.00$ |
| 5 vs. 4 | $p < .001$ | $p < .001$ | $p < .001$ | $p < .001$ | $p = .61$ | $p = .78$ |
| 6 vs. 4 | $p < .001$ | $p < .001$ | $p < .001$ | $p < .001$ | $p = 1.00$ | $p = .98$ |
| 6 vs. 5 | $p = .99$ | $p < .001$ | $p < .001$ | $p < .001$ | $p = .72$ | $p = .37$ |

## Model Generalizability

**Internal generalizability**. When performing unsupervised cluster analyses, it is important to know whether the cluster results generalize (e.g., Research Question 2a). In the current research, a bootstrapping and replication methodology was adopted to see if the observed clusters generalize meaningfully to unseen data (Dalton, Ballarin, & Brun, 2009; Everitt, Landau, Leese, & Stahl, 2011). First, the internal generalizability was evaluated for the six-cluster model from the Traditional CSCL dataset. Specifically, a bootstrapping approach was used to assess the "prediction strength" of the training data. The prediction strength measure assesses how many groups can be predicted from the data, and how well (Tibshirani & Walther, 2005). Following the prediction strength assessment, a replication model was used to evaluate whether the training data cluster centroids can predict the ones in the testing data. If the six-cluster structure found using k-means clustering is appropriate for the Traditional CSCL data, then the prediction for the test dataset, and a clustering solution created independently for the test dataset, should match closely.

The prediction strength of the training data was explored using the *clusterboot* function in the R package fpc (Hennig, 2015). This approach uses a bootstrap resampling scheme to evaluate the prediction strength of a given cluster. The algorithm uses the *Jaccard coefficient*, a similarity measure between sets. The Jaccard similarity between two sets Y and X is the ratio of the number of elements in the intersection of Y and X over the number of elements in the union of Y and X. The cluster prediction strength and stability of each cluster in the original six-cluster model is the mean value of its Jaccard coefficient over all the bootstrap iterations. As a rule of thumb, clusters with a value less than 0.6 should be considered unstable. Values between 0.6 and 0.75 indicate that the cluster is measuring a pattern in the data, but there is not high certainty about which points should be clustered together. Clusters with values above about 0.85 can be considered highly stable and have high prediction strength (Zumel, Mount, & Porzak, 2014). The prediction strength of the Traditional CSCL training data was evaluated using 100 bootstrap resampling iterations.

The final cluster pattern produced by the 100 bootstrap resampling iterations for the six-cluster model are reported in Figure 5. As seen in the Figure 5, the observed pattern was identical to the original six-cluster model, albeit with a different ordering of the clusters. The ordering of clusters in the k-means algorithm is arbitrary so the pattern of the GCA variables within each cluster is of most importance. The Jaccard's similarity values showed very strong prediction for all six clusters with .96, .95, .91, .96, .91, and .96 for clusters 1-6, respectively.

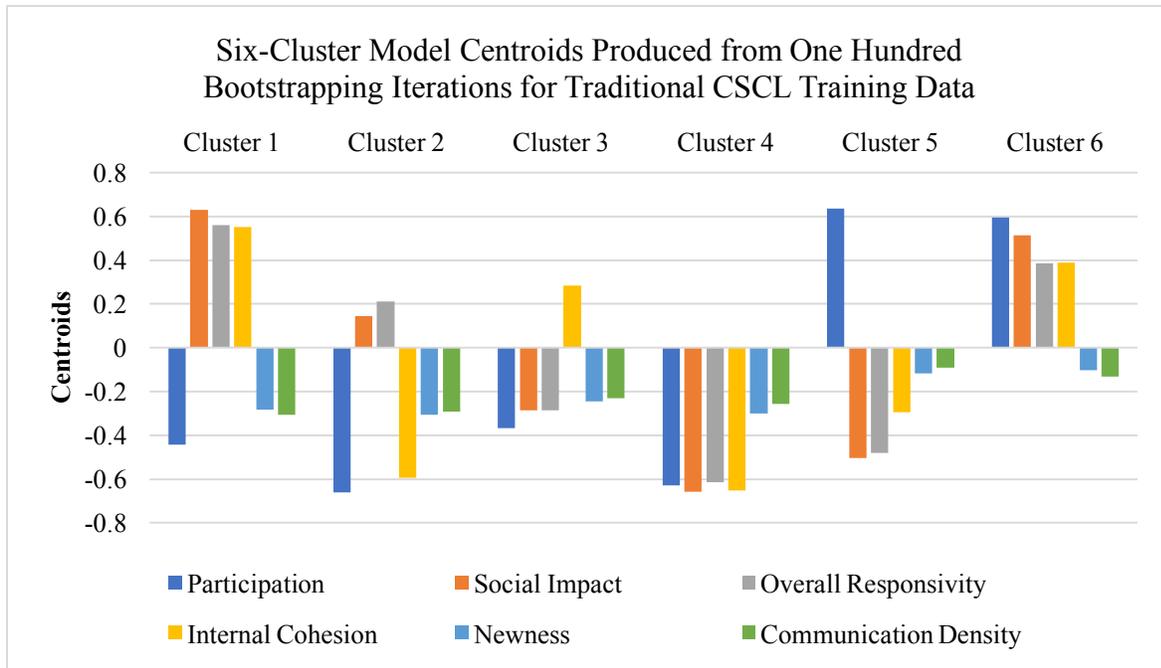

Figure 5. This figure shows the final six-cluster pattern produced by the 100 bootstrap resampling iterations Traditional CSCL training data, which was identical to the original k-means six-cluster model pattern depicted in Figure 4.

The next analyses focus on evaluating the generalizability of the observed clusters in the training data to the testing data. First, six-cluster k-means analyses were performed on the held out Traditional CSCL test data ($N$= 136). Descriptive statistics for the test data GCA variables are reported in the supplementary material. The centroids for the six-cluster k-means solution for the Traditional CSCL test data are illustrated in Figure 6. The observed pattern of the six-cluster solution for the testing data appears, at least visually, similar to the one observed on the training data.

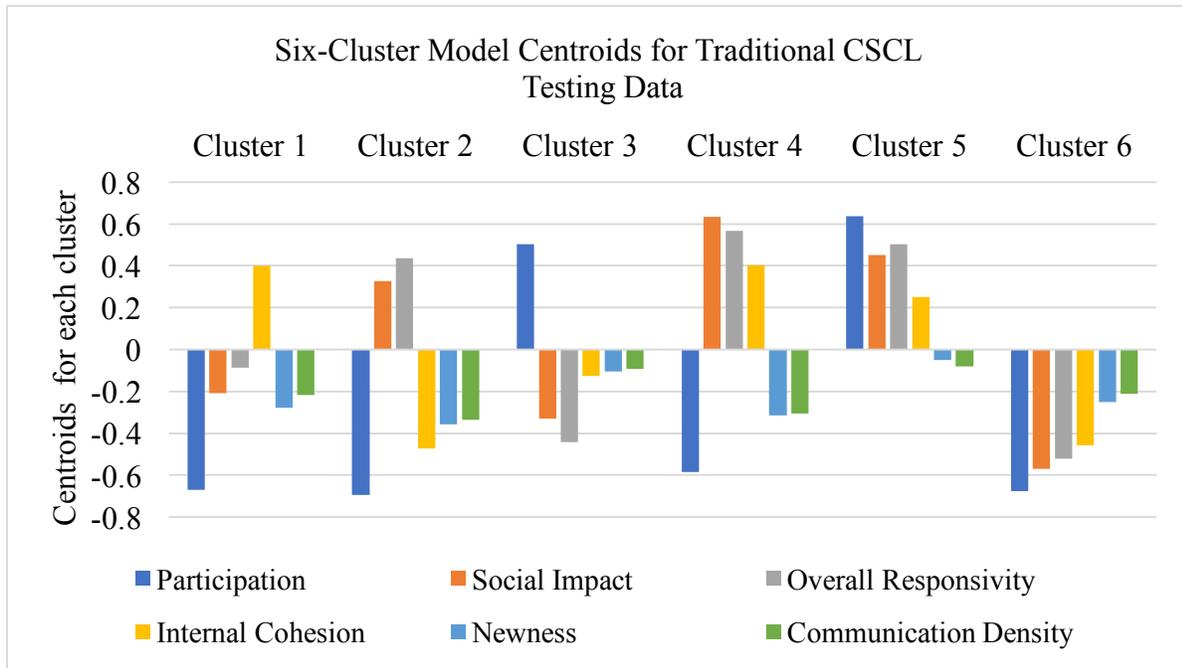

Figure 6. Traditional CSCL testing data centroids for the six-cluster solution across the GCA variables.

The next analyses focus on quantifying the observed overlap between the testing and training cluster analyses. Specifically, the cluster centers from the training dataset were used to predict the clusters in the test data for the six-cluster model. This analysis was performed using the *cl_predict* function in the R clue package (Hornik & Böhm, 2016). Cross-tabulation of the predicted and actual cluster assignments for the Traditional CSCL testing data set are reported in Table 6. The rows in the table correspond to the clusters specified by the k-means clustering on the testing data and the columns correspond to the predicted cluster membership by the training data. In a perfect prediction, large values would lie along the diagonal, with zeroes off the diagonal; that would indicate that all samples that belong to cluster 1 were predicted by the training data as belonging to cluster 1, and so forth. The form of this table can give us considerable insight into which clusters are reliably predicted. It can also show which groups are likely to be confused and which types of misclassification are more common than others. However, in this case we observed an almost perfect prediction in the six-cluster model, with few exceptions.

**Table 6** Cross-tabulation of the Predicted and Actual Cluster Assignments for the Six-Cluster Model on Traditional CSCL Testing Data Set

| Testing Clusters | Training Predicted Clusters | | | | | |
|---|---|---|---|---|---|---|
| | Cluster 1 | Cluster 2 | Cluster 3 | Cluster 4 | Cluster 5 | Cluster 6 |
| Cluster 1 | 32 | 0 | 0 | 0 | 0 | 0 |
| Cluster 2 | 2 | 29 | 0 | 0 | 0 | 0 |
| Cluster 3 | 0 | 0 | 15 | 2 | 1 | 0 |
| Cluster 4 | 0 | 0 | 0 | 18 | 0 | 0 |
| Cluster 5 | 4 | 0 | 0 | 1 | 13 | 0 |
| Cluster 6 | 0 | 0 | 0 | 0 | 0 | 19 |

Two measures were used to evaluate the predictive accuracy of the six-cluster model on the Traditional CSCL training clusters: Adjusted Rand Index (ARI) and a measure of effect size (Cramer $V$) for the cluster cross-tabulation. ARI computes the proportion, of the total of $\binom{n}{2}$ object pairs, that agree; that is, are either (i) in the same cluster according to partition 1 and the same cluster according to partition 2 or (ii) in different clusters according to 1 and in different clusters according to 2. The ARI addresses some of the limitations of the original rand index by providing a conservative measure which penalizes for any randomness in the overlap (Hubert & Arabie, 1985). The ARI was calculated between: (a) the test data clustering membership and (b) the predicted cluster membership given by the training data. The predictive accuracy of the training data is considered good if it is highly similar to the actual testing data cluster membership. The degree of association between the membership assignments of the predicted and actual cluster solutions was ARI = 0.84 for the six-cluster model. ARI values range from 0 to 1, with higher index values indicating more agreement between sets. The measure of effect size for the cross-tabulation revealed Cramer $V$ = 0.92, which is considered very strong association (Kotrlik, Williams, & Jabor, 2011). Given these results, the six-cluster solution was judged to be robust and well supported by the data.

A similar replication approach was adopted to evaluate the generalizability within the SMOC and Land Science datasets. Descriptive statistics for the GCA measures in the SMOC training ($N$ = 9,463)/ testing ($N$ = 2,378) and Land Science training ($N$ = 2,837)/ testing ($N$ =

695) data sets are presented in Table 7. First, a six-cluster model was constructed on the SMOC and Land Science training datasets. The pattern of the six-cluster models are depicted in Figure 7 for the SMOC training dataset, and Figure 8 for the Land Science training dataset.

**Table 7** Descriptive Statistics for GCA Measures in the SMOC & Land Science Training and Testing Data Sets

| Measure | Min | | Med | | $M$ | | $SD$ | | Max | |
|---|---|---|---|---|---|---|---|---|---|---|
| **SMOC** | Train | Test | Train | Test | Train | Test | Train | Test | Train | Test |
| Participation | -0.44 | -0.49 | 0.00 | 0.00 | 0.00 | 0.00 | 0.11 | 0.11 | 0.45 | 0.42 |
| Social Impact | -0.14 | -0.05 | 0.15 | 0.15 | 0.16 | 0.16 | 0.10 | 0.10 | 1.00 | 1.00 |
| Overall Responsivity | -0.30 | -0.04 | 0.15 | 0.15 | 0.16 | 0.16 | 0.11 | 0.11 | 1.00 | 1.00 |
| Internal Cohesion | -0.43 | -0.17 | 0.12 | 0.12 | 0.13 | 0.14 | 0.12 | 0.12 | 1.00 | 1.00 |
| Newness | 0.00 | 0.00 | 0.65 | 0.65 | 0.84 | 0.83 | 0.83 | 0.76 | 17.39 | 7.73 |
| Communication Density | 0.00 | 0.00 | 0.19 | 0.19 | 0.26 | 0.26 | 0.30 | 0.26 | 10.56 | 3.32 |
| **Land Science** | | | | | | | | | | |
| Participation | -0.50 | -0.49 | -0.01 | -0.03 | 0.00 | 0.00 | 0.14 | 0.15 | 0.78 | 0.49 |
| Social Impact | -0.10 | -0.05 | 0.12 | 0.12 | 0.13 | 0.12 | 0.09 | 0.08 | 0.90 | 0.74 |
| Overall Responsivity | -0.12 | -0.04 | 0.11 | 0.11 | 0.13 | 0.12 | 0.10 | 0.09 | 1.00 | 1.00 |
| Internal Cohesion | -0.21 | -0.17 | 0.11 | 0.11 | 0.13 | 0.12 | 0.13 | 0.12 | 1.00 | 1.00 |
| Newness | 0.00 | 0.00 | 0.60 | 0.59 | 1.10 | 1.11 | 2.33 | 2.15 | 70.27 | 27.39 |
| Communication Density | 0.00 | 0.00 | 0.18 | 0.18 | 0.38 | 0.36 | 0.94 | 0.72 | 31.27 | 10.45 |

*Note*. Mean (**$M$**). Standard deviation (**$SD$**).

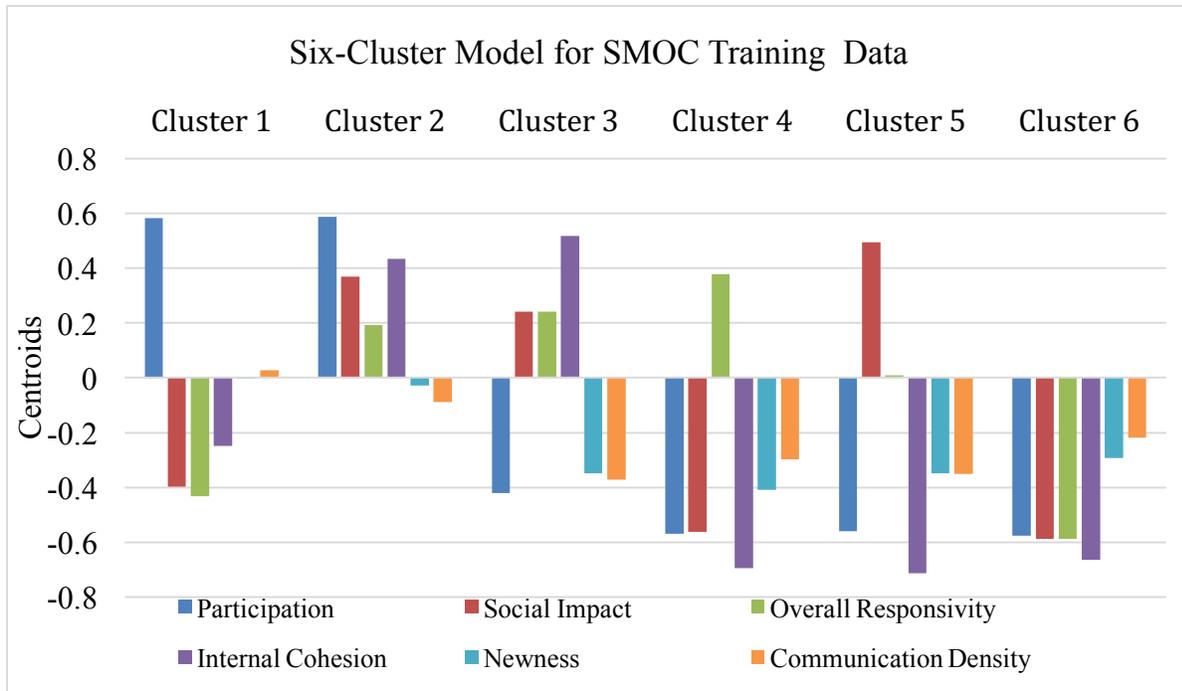

*Figure 7.* SMOC training data centroids for the six-cluster solution across the GCA variables.

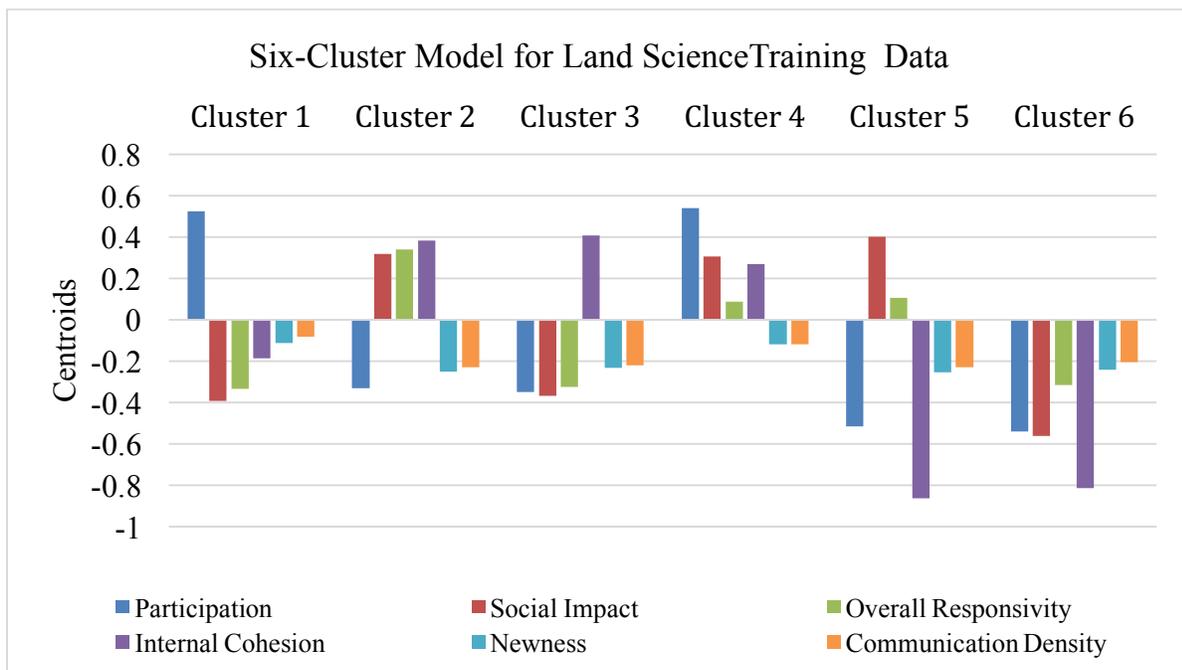

*Figure 8.* Land Science training data centroids for the six-cluster solution across the GCA variables.

The analysis proceeded by evaluating the internal generalizability for the SMOC and Land Science datasets separately. This analysis was performed by using the cluster centroids from the SMOC and Land Science training datasets to predict the clusters in the test data for the six-cluster model. These analyses were also performed using the *cl_predict* function in the R clue package (Hornik & Böhm, 2016). Cross-tabulation of the predicted and actual cluster assignments for the SMOC and Land Science testing dataset are reported in Tables 8 and Table 9, respectively. We see from these tables that there appears to be good agreement for the predicted cluster assignments in the six-cluster models. We can quantify the agreement using the ARI and Cramer $V$ provided by the flexclust package. For the SMOC dataset and Land Science dataset, ARI = 90 and ARI = 86, respectively. Again, the ARI values range from 0 to 1, with higher index values indicating more agreement between sets. This suggest there was the six-cluster model exhibited slightly higher predictive agreement between the training and testing data cluster assignments for the SMOC dataset when compared to the Land Science. However, both the SMOC and Land Science datasets had high effect sizes with Cramer $V$ = 95 and Cramer $V$ = 92, respectively. Taken together, the six-cluster solutions were judged to be supported by both the SMOC collaborative interaction data and Land Science collaborative problem-solving data, with the six-cluster model being only minimally better internal generalizability.

**Table 8** Cross-tabulation of the Six-Cluster Model Predicted and Actual Cluster Assignments for the SMOC Testing Data Set

| Testing Clusters | Predicted Clusters | | | | | |
|---|---|---|---|---|---|---|
| | Cluster 1 | Cluster 2 | Cluster 3 | Cluster 4 | Cluster 3 | Cluster 5 |
| Cluster 1 | 517 | 17 | 4 | 0 | 1 | 15 |
| Cluster 2 | 0 | 469 | 14 | 0 | 0 | 0 |
| Cluster 3 | 0 | 5 | 475 | 1 | 0 | 10 |
| Cluster 4 | 1 | 0 | 1 | 208 | 0 | 4 |
| Cluster 5 | 0 | 0 | 6 | 6 | 198 | 0 |
| Cluster 6 | 1 | 0 | 0 | 3 | 7 | 415 |

**Table 9** Cross-tabulation of the Six-Cluster Model Predicted and Actual Cluster Assignments for the Land Science Testing Data Set

| Testing Clusters | Predicted Clusters | | | | | |
|---|---|---|---|---|---|---|
| | Cluster 1 | Cluster 2 | Cluster 3 | Cluster 4 | Cluster 3 | Cluster 5 |
| Cluster 1 | 137 | 0 | 0 | 0 | 1 | 1 |
| Cluster 2 | 0 | 90 | 3 | 9 | 4 | 0 |
| Cluster 3 | 1 | 12 | 81 | 0 | 0 | 0 |
| Cluster 4 | 11 | 0 | 2 | 106 | 0 | 0 |
| Cluster 5 | 0 | 0 | 0 | 0 | 98 | 0 |
| Cluster 6 | 0 | 0 | 0 | 0 | 1 | 138 |

**External generalizability**. The practice of predictive modeling defines the process of developing a model in a way that we can understand and quantify the model's prediction accuracy on future, yet-to-be-seen data (Kuhn & Johnson, 2013). The previous analyses provided confidence in the six-cluster models' ability to generalize to unseen data within the same data set. However, the ultimate goal is to evaluate how well the identified student roles (i.e., clusters) are representative of interaction patterns across various types of collaborative interactions. This step is critical because the robustness and accuracy of the models across datasets will determine the usefulness of the GCA for broader research applications. Thus, the next analyses assess the generalizability of these clusters across the three collaborative interaction datasets (i.e., Research Question 2b). Specifically, the clusters centers from each data set were used to predict the clusters in the other training datasets, wherein all possible combinations were evaluated. Again, two measures were used to evaluate the predictive accuracy of clusters: ARI, and a measure of effect size, Cramer V, for their cross-tabulation. Table 10 shows the ARI and Cramer $V$ results for the computed cross-tabulation evaluations of the six-cluster models. The columns in Table 10 correspond to the predictor dataset, while the rows correspond to the predicted data set.

The first item to take away from Table 10 is that, the predictive accuracy (ARI) is lower for all datasets compared to the previously reported internal generalization evaluations. This overall drop in predictive accuracy is to be expected with evaluating external data. While the accuracy is lower than the internal evaluations, the ARI results are still quite high for the majority of the predictions. Specifically, we see the SMOC dataset has the lowest agreement predicting clusters in the Traditional CSCL and Land Science. However, Land Science had the highest agreement with predicting the Traditional CSCL, and was on par with the Traditional CSCL when predicting the SMOC dataset.

**Table 10** ARI and Cramer V Results for the Cluster Model Computed Cross-Tabulation Tables

| Model | W3 Training | | SMOC Training | | Land Science Training | |
|---|---|---|---|---|---|---|
| Six-Cluster Model | ARI | Cramer $V$ | ARI | Cramer $V$ | ARI | Cramer $V$ |
| W3 Training Data | -- | -- | .66 | .89 | .76 | .86 |
| SMOC Training Data | .70 | .78 | -- | -- | .69 | .79 |
| Land Science Training Data | .69 | .83 | .66 | .78 | -- | -- |

*Note.* -- indicates previously reported internal generalization evaluations, which are not reported here to avoid redundancy.

**Discussion**

We explored the extent to which the characteristics of collaborative interaction discourse, as captured by the GCA, diagnostically reveal the social roles students occupy, and if the observed patterns are both robust and generalize. The findings present some methodological, conceptual and practical implications for the group interaction research, educational data mining and learning analytics communities. The GCA represents a novel methodological contribution, capable of identifying distinct patterns of interaction representative of the social roles students occupy in collaborative interactions. The natural language metrics that make up the GCA provide a mechanism to operationalize such roles, and provide a view on how they are constructed and maintained through the sociocognitive processes within an interaction. We expect the GCA to provide a more objective, domain independent, and deeper exploration of the micro-level inter- and intra-personal patterns associated with social roles. Moreover, as the methodology is readily automated, substantially larger corpora can be analyzed with the GCA than is practical when human judgements are required to annotate the data.

The current research has built upon the framework of Strijbos and De Laat (2010) by adding several new dimensions of interaction. The GCA measures revealed behavioral and communication patterns described by Strijbos and De Laat's (2010), and in this respect the current research has replicated some of their results. Interestingly, however, our analyses uncovered social roles that do not entirely overlap with those observed in their work. In this respect, we have been able to build upon their results and to provide insights beyond what their framework revealed. The identification of these additional roles may serve as a useful conceptual addition for future research focusing on the social roles within multi-party communication. For instance, only one role, the Over-rider, appeared to overlap in in the six-cluster model for the Traditional CSCL dataset. However, the other roles did not appear to align with the labels suggested Strijbos and De Laat's (2010) framework. This is likely due to the fact that the GCA includes more and different dimensions than are represented in their framework.

The identified social roles (i.e., clusters) underwent stringent evaluation and validation assessments: internal criteria, stability, coherence and generalizability. Internal criteria measures evaluated the extent to which the clustering "fits" the training data. The findings suggested that the six-cluster model performed well across the three internal criteria measures. The stability measures captured the extent to which the clusters remain intact even when the dataset is perturbed through the loss of a dimension. Here, we saw that the six-cluster solution is quite stable. The cluster coherence allowed us to test whether the GCA dimensions differed significantly from one role to the other. This coherence evaluation showed that the six-cluster model exhibited nice separation across the GCA measures. Finally, the cluster models were inspected for their ability to generalize both within and across the three datasets. These internal and external generalization tests provide us with confidence in the robustness for the identified roles. Given the extent of these evaluations, we feel that the roles identified can be considered as robust and stable constructs in the space of small group interactions, and that the GCA measures capture the critical socio-cognitive processes necessary for identifying such roles.

**Student Roles and Learning**

Unlike the internal criteria explored in the above section, external criteria are independent of the way the clusters are obtained. External cluster validation can be explored by either comparing the cluster solutions to some "known" categories or by comparing them to meaningful external variables, i.e. variables not used in the cluster analysis (Antonenko, Toy, & Niederhauser, 2012). Furthermore, the practical impact of the identified social roles may be felt at multiple of levels of granularity, and we therefore must test for their impact at multiple levels. In the current research the usefulness of identifying learners' roles in collaborative learning is explored through two analyses of the data: (a) the influence of student roles on individual student performance, and (b) the influence of student roles on overall group performance (Research Questions 3a & 3b).

The multi-level investigation conducted in the current research also addresses a frequently noted limitation found in collaborative learning research. CSCL researchers encounter issues regarding the differing units of analysis in their datasets (Janssen, Erkens, Kirschner, & Kanselaar, 2011). That is, collaborative interactions can be analyzed at the level of the group, the individual student, and of each student-student interaction. For example, in the current research, some variables of interest are measured at the individual learner and interaction levels (e.g., student learning gains, participation, internal cohesion, social impact, overall responsivity, newness, communication density, and social roles identified by the cluster analysis), whereas other variables are measured at the group level (e.g., group diversity, group composition, and group performance). Several researchers have emphasized the need to conduct more rigorous, multi-level research (Cress, 2008; Bram De Wever, Van Keer, Schellens, & Valcke, 2007; Stahl, 2005; Suthers, 2006b). However, collaborative learning studies usually focus on only one of these levels (Stahl, 2013). As a result, there is little consideration given as to how these levels are connected, despite it being well-recognized that such connections are crucially important to both understanding and orchestrating learning in collaborative learning environments (Stahl, 2013). To avoid this problem, a series of models were constructed to explore both the influence of group-level constructs on individual student-level learning, as well as individual student-level constructs on group performance.

A student-level performance score was obtained for each student by calculating their proportional learning gains, formulated as in (Hake, 1998):

$$\frac{(\%PostTest - \%PreTest)}{(100 - \%PreTest)} \tag{35}$$

Correlations between learning gains and the six GCA variables in the Traditional CSCL dataset are reported in Table 11.

**Table 11** Correlations between Learning and GCA Variables in the Traditional CSCL Dataset

|  | Learning Gains | Participation | Social Impact | Overall Responsivity | Internal Cohesion | Newness |
|---|---|---|---|---|---|---|
| Participation | 0.10** |  |  |  |  |  |
| Social Impact | 0.10* | 0.07 |  |  |  |  |
| Overall Responsivity | 0.10* | -0.01 | 0.69*** |  |  |  |
| Internal Cohesion | 0.13*** | 0.21*** | 0.57*** | 0.52*** |  |  |
| Newness | 0.06 | 0.62*** | 0.05 | -0.03 | 0.11** |  |
| Communication Density | 0.04 | 0.54*** | -0.11*** | -0.18*** | -0.05 | 0.91*** |

*Note.* $*** p < .001. ** p < .01. * p < .05.$

A mixed-effects modeling methodology was adopted for these analyses due to the nested structure of the data (e.g., students within groups) (Pinheiro & Bates, 2000). Mixed-effects models include a combination of fixed and random effects, and can be used to assess the influence of the fixed effects on dependent variables after accounting for the random effects. Multilevel modelling handles the hierarchical nesting, interdependency, and unit of analysis problems that are inherent in collaborative learning data. They are the most appropriate technique for investigating data in CSCL-environments (De Wever et al., 2007; Janssen et al., 2011). Table 12 provides an overview of the mixed-effects models used to explore such potential multi-level effects across the six-cluster solution.

**Table 12** Overview of Mixed-Effects Models Exploring Learning across the Six Cluster Solution

| Model Number | Dependent Variable | Level of Dependent Variable | Independent Variable | Level of Independent Variable | Random Variable(s) |
|---|---|---|---|---|---|
| 1 | Learning Gains | Student | Social Roles | Student | Student Nested in Group |
| 2 | Learning Gains | Student | Role Diversity | Group | Student Nested in Group |
| 3 | Performance | Group | Role Diversity | Group | Group |
| 4-6 | Learning Gains | Student | Proportional Occurrence Roles | Group | Student Nested in Group |
| 7-9 | Performance | Group | Proportional Occurrence Roles | Group | Group |

In addition to constructing the fixed effects models, null models with the random effects (the student nested in the group or the group) but no fixed effects were also constructed. A comparison of the null random-effects only model with the fixed-effect models allows us to determine whether social roles and communication patterns predict student and group performance above and beyond the variance attributed to individual students or groups. The Akaike Information Criterion (AIC), Log Likelihood (LL) and a likelihood ratio test were all used to evaluate the overall fit of the models. Additionally, the effect sizes for each model were estimated using a pseudo $R^2$ method, as suggested by Nakagawa and Schielzeth (Nakagawa & Schielzeth, 2013). For mixed-effects models, $R^2$ can be divided into two parts: marginal ($R^2_m$) and conditional ($R^2_c$). Marginal $R^2$ is associated with variance explained by fixed factors, whereas conditional $R^2$ can be interpreted as the variance explained by the entire model, namely random and fixed factors. Both the marginal and conditional parts convey unique and relevant information regarding the model fit and variance explained. The nlme package in R (Pinheiro et al., 2016) was used to perform all the required computations. All analyses are on the Traditional

CSCL dataset because it was the base corpus for the cluster analyses and it has the most consistent individual and group performance measures.

**Influence of Student Roles on Individual Student Performance** In order to evaluate the effects of roles purely at the individual level, two linear mixed-effects models were compared: (a) model 1 from Table 12, with learning gains as the dependent variable, social roles as independent variables, and student nested within group as the random effects, and (b) the null model with random effects only and no fixed effects. The likelihood ratio tests indicated that the six-role model with $\chi^2(5) = 11.55$, $p = .04$, $R^2_m = .02$, $R^2_c = .95$ yielded a significantly better fit than the null model. A number of conclusions can be drawn from these statistics. Firstly, that the roles in the six-cluster model were able to add significantly to the prediction of the learners' performance beyond the variance attributed to student and group membership. Second, social roles, individual participant, and group features explained about 95% of the predictable variance, with 2% of the variance being accounted for by the social roles.

The social roles that were predictive of individual student learning performance for the six-cluster model are presented in Table 13. The reference group was the Driver role, meaning that the learning gains for the other roles are compared against the Driver reference group. Four of the six roles exhibit significant differences in student learning gains, as compared to the Driver role. Here we see learners who took on more socially responsible, collaborative roles, such as the Driver, performed significantly better than students who occupied the less socially engaged roles, like Lurker, and Over-rider. There was no significant difference between the performance of the Driver and Task-Leader or Socially Detached, suggesting these are the more successful roles in terms of student learning gains.

**Table 13** Descriptive Statistics for Student Learning Gains Across Six Roles and Mixed-Effects Model Coefficients for Predicting Differences in Individual Student Performance Across Clusters

| Role | Six-Cluster Model | | | |
|------|------|------|------|------|
| | *M* | *SD* | β | *SE* |
| Driver | 0.21 | 0.89 | 0.21** | 0.07 |
| Over-rider | 0.02 | 0.88 | -0.19* | 0.10 |
| Lurker | -0.11 | 0.79 | -0.32** | 0.11 |
| Follower | -0.08 | 0.92 | -0.29** | 0.12 |
| Socially Detached | 0.03 | 0.83 | -0.18 | 0.11 |
| Task-Leader | 0.09 | 0.84 | -0.12 | 0.10 |

*Note. * $p < .05$. ** $p < .01$. *** $p < .001$. Mean (M). Standard deviation (SD). Fixed effect coefficient (β). Standard error (SE).*

It is important to note that the observed difference in learning gains across the social roles is not a result of the students simply being more prolific because the Task-Leaders and Socially Detached learners performed on par with the Drivers, but were among of the lower participators in the group. The profile for the Socially Detached learners showed mid-range values for responsivity and social impact, compared to their internal cohesion scores. However, the Task-Leaders profile illustrated that when they did make contributions it was very responsive to the other group members (i.e., high overall responsivity), as well as being semantically connected with their previous contributions (i.e., high internal cohesion). Furthermore, their contributions were seen as relevant by their peers (i.e., high social impact). These findings reflected a more substantive difference in social awareness and engagement for the Drivers and Task-Leaders, compared to the Over-riders, beyond the surface level mechanism of simply participating often. Taken together, these results show that the identified roles are externally valid; not just due to a significant relationship to the external measure of learning, but also because the we can make theoretically meaningful predictions from the roles associated characteristic behaviors.

**Incorporating Group Level Measures**

Two groups of models were constructed to assess the influence of group composition on group performance and individual student learning gains. The first set of models (i.e., models 2 and 3 from Table 12) assessed the influence of group role diversity on student learning gains and group performance. The second set of models (i.e., models 4-6 and 7-9 from Table 12) dove deeper to explore the influence of group compositions, as measured by the proportional occurrence of each of the roles, on student learning gains and group performance. As a reminder, group performance was operationally defined as the amount of topic-relevant discussion during the collaborative interaction (equation 34).

The proportional occurrence (frequency) of each role within any group can be a helpful measure in determining group composition. For a group G, it can be formally defined as:

$$\hat{p}_G(r) = \frac{\#\ users\ of\ role\ r\ in\ G}{size\ of\ G}$$

Group compositions was operationalized using a measure of role diversity, based on entropy. Entropy is a measure at the core of information theory quantifying the amount of "surprise" possible in a probability distribution. At the extremes, entropy ranges from values of 0 for distributions where a single outcome is always the case (i.e. $P(X = x) = 1.0$), to a maximum value when the probability of all outcomes is equal (i.e. a uniform distribution). The entropy of roles in a group will then be 0 for groups where all participants take on the same role, and greater for groups with a greater diversity of roles. Role diversity for a group is calculated as:

$$H(G) = -\sum_{r \in Roles} \hat{p}_G(r) \cdot \log\left(\hat{p}_G(r)\right)$$

Correlations between group performance, student learning gains, role diversity, and the proportional occurrences of each role are reported in Table 14. No relationship was observed between student learning gains and group performance, so this was not probed further. Quite small relationships were observed between role diversity ($M = 1.04$, $SD = .26$) with student learning gains and group performance. However, when these relationships were further explored, the likelihood ratio tests indicated that the full diversity models for student learning gains and group performance did not yield a significantly better fit than the null model with $\chi^2(1) = .39$, $p = .52$, $R^2_m = .001$, $R^2_c = .96$, and $\chi^2(1) = .26$, $p = .62$, $R^2_m = .002$, $R^2_c = .88$, respectively.

**Table 14** Correlations between Student Learning Gains, Group Performance, Role Diversity and the Proportional Occurrence of Six Roles

| | Student Level | Group Level Measures | | | | | | |
|---|---|---|---|---|---|---|---|---|
| Measure | Learning Gains | Group Performance | Diversity | Prop. Over-rider | Prop. Driver | Prop. Follower | Prop. Lurker | Prop. Socially Detached |
| Group Performance | 0.00 | | | | | | | |
| Diversity | -0.02 | -0.03 | | | | | | |
| Prop. Over-rider | -0.03 | -0.28*** | 0.03 | | | | | |
| Prop. Driver | 0.03 | 0.28*** | -0.12*** | -0.77*** | | | | |
| Prop. Follower | -0.01 | 0.02 | 0.12*** | -0.31*** | 0.29*** | | | |
| Prop. Lurker | -0.05 | -0.28*** | -0.04 | 0.47*** | -0.49*** | -0.46*** | | |
| Prop. Socially Detached | -0.01 | -0.13*** | 0.23*** | 0.16*** | -0.43*** | -0.29*** | 0.07 | |
| Prop. Task-Leader | 0.05 | 0.32*** | -0.16*** | -0.47*** | 0.28*** | -0.11** | -0.52*** | -0.37*** |

*Note.* *** $p < .001$. ** $p < .01$. * $p < .05$.

The second set of analyses involved a more fine-grained investigation of the influence of (the proportional occurrence of) positive and negative roles on student learning gains and group performance. Six linear mixed-effects models were constructed, where three models were student-level (i.e., models 4-6 from Table 12), and three models were group-level (i.e., models 7-9 from Table 12). Particularly, we constructed a *productive roles model* with the proportional occurrence of Drivers, Task-Leaders, and Socially Detached learners as the independent variable, and an *unproductive roles model* with the proportional occurrence of Over-riders, Followers and Lurkers as the independent variable. Null models were constructed for both the student and group level analyses. For the six models below, the first three models had student learning gains as the dependent variable, whereas the next three had group performance as the dependent variable.

For the student level analyses, the likelihood ratio tests indicated that neither the *productive role* model nor the *unproductive role* model yielded a significantly better fit than the null model with $\chi^2(3) = 2.62$, $p = .45$, $R^2_m = .004$, $R^2_c = .96$, and $\chi^2(3) = 2.75$, $p = .43$, $R^2_m = .004$, $R^2_c = .96$. When we combine this with the previous finding that social role does influence individual learning, this suggests that it is less important that a person has productive roles in one's group than it is that the individual is enacting a productive role.

For the group level analysis, the likelihood ratio tests indicated that that both the *productive roles model* and the *unproductive roles model* yielded a significantly better fit than the null model with $\chi^2(3) = 23.62$, $p < .0001$, $R^2_m = .15$, $R^2_c = .90$, and $\chi^2(3) = 20.92$ $p < .001$, $R^2_m = .13$, $R^2_c = .89$, respectively. Several conclusions can be drawn from this model comparison. First, that the proportional occurrence of productive and unproductive roles, were able to significantly improve predictions of the group performance above and beyond the variance attributed to group. Secondly, that for all models, the proportional occurrence of different social roles combined with group features explained about 89% of the predictable variance in group performance, with 28% of the variance being accounted for by the proportional occurrence of different social roles. Table 15 shows the social roles that were predictive of group performance for both the *productive roles model* and the *unproductive roles model*.

As shown in Table 15, groups with greater proportions of learners who took on more socially responsible, collaborative roles (namely Driver and Task-Leader) performed significantly better than groups with greater proportions of less socially engaged roles (Lurker and Over-rider). These findings mirror the pattern that was observed for individual student learning and social roles.

**Table 15** Descriptive Statistics for Group Performance Across Six Roles and Mixed-Effects Model Coefficients for Predicting the Influence of Productive and Unproductive Roles on Group Performance

| Role | **Productive Roles Model** | | | | Role | **Unproductive Roles Model** | | | |
|---|---|---|---|---|---|---|---|---|---|
| | *M* | *SD* | β | *SE* | | *M* | *SD* | β | *SE* |
| Prop. of Driver | 0.27 | 1.05 | 1.15** | 0.41 | Prop. of Over-Rider | -0.27 | 0.92 | -1.05* | 0.46 |
| Prop. of Socially Detached | -0.18 | 0.79 | 0.42 | 0.52 | Prop. of Follower | 0.03 | 0.94 | -1.02* | 0.55 |
| Prop. of Task-Leader | 0.37 | 1.04 | 1.27** | 0.39 | Prop. of Lurker | -0.32 | 0.94 | -1.42* | 0.52 |

*Note. N = 148. * p < .06. ** p < .01. Mean (M). Standard deviation (SD). Fixed effect coefficient (β). Standard error (SE).*

## Discussion

We investigated whether the social roles (Driver, Task-Leader, Lurker, and Over-rider, Socially Detached, and Follower) were meaningfully related to both student learning gains and group performance. Overall, the results suggest that the roles that learners occupy influences their learning, and that the occurrence of certain roles within a group interaction can result in different outcomes for that group. Furthermore, we established the connection between the individual-level and group-level outcomes is affected by the same productive or un-productive roles. Taken together, these discoveries show that not only are the identified roles related to learning and to collaborative success, but that this relationship is theoretically meaningful, which provides external validity.

For the individual student learning models, we saw that socially engaged roles, like Driver, significantly outperformed less participatory roles, like Lurkers. This finding might be expected. However, other findings emerged that were less intuitive. For instance, we found that Task-Leaders and Socially Detached leaners performed comparably well with the Drivers (although not quite as high), but were among of the lower participators in the group. This suggests the difference in learning gains across the social roles is not simply a result of the students being more prolific. Clearly engagement or mastery with the material can be manifested not only in greater quantity, but also quality of participation. The Task-Leaders were highly responsive, and had high social impact and internal cohesion, but lower scores for newness and communication density. However, the most defining feature of the Socially Detached learners was their high internal cohesion because they exhibited relatively mediocre scores across the other GCA measures. Something interesting starts to emerge when these profiles are juxtaposed with the Over-riders. Over-riders were the highest participators, but had lower learning gains, responsivity to peers, social impact, and mediocre internal cohesion. Together, this highlights the potency of internal cohesion, and being even mildly socially aware and engaged with other group members. More than simply talking a lot, the intra- and inter-personal dynamics appear to be major factors in how students perform in groups.

The influence of the roles on group performance was also investigated. We started by looking at the influence of the overall diversity of roles on group performance. Here, we were interested in seeing if groups that are comprised of, for example, six different roles performed better than those that were comprised of all Task-Leaders. This was motivated by the group interaction literature, which suggests that diversity can be a major contributor to the successfulness of collaborative interactions. The findings for diversity in the literature have explored several different types of diversity, including personality, prior knowledge, gender, and other individual traits (Barron, 2003; Fuchs, Fuchs, Hamlett, & Karns, 1998). These analyses did not suggest any significant influence of role diversity on student or group performance, suggesting, perhaps, that diversity in roles is not an important type of diversity.

We then dove deeper into an investigation of group composition as given by the proportional occurrence of each role. The findings here were considerably more promising, and largely mirrored those found for the individual students, with a few exceptions. In particular, we observed that the presence of Socially Detached learners within a group did not significantly influence the group performance. This is most likely because, while they may be successful students individually, they do not engage meaningfully with their peers, and so have little impact on the group. These findings have implications optimal group composition, suggesting that groups should not simply be comprised of high participating members, but include a combination of both low and high participators. However, what is perhaps even more important is that the group include members that are both aware of the social climate of the group interaction and invested in the collaborative outcome.

Another difference between the influence of roles on groups and individual performance pertains to the effect sizes. The influence of roles within a group appears to have a more potent influence on group performance (explaining 26%-28% of the variance) than does the influence of taking on a particular role on student performance (explaining only 2% of the variance). This illustrates the substantial impact that even a few members can have on a group, and the importance of diligent orchestration for optimal group composition.

## Conclusion

A primary objective of this research was to propose and validate a novel automated methodology, *Group Communication Analysis* (GCA), for detecting emergent roles in group interactions. The GCA applies automated computational linguistic techniques to the sequential interactions of online collaborative interactions. The GCA involves computing six distinct measures of sociocognitive interaction patterns (i.e., Participation, Overall Responsivity, Social Impact, Internal Cohesion, Communication Density, and Sharing of New Information). The automated natural language metrics that make up the GCA provide a new and useful view on how roles are constructed and maintained in collaborative interactions.

There are some notable limitations to the variables selected for inclusion in the GCA. Particularly, the current research focused only on sociocognitive variables, however, there are several other collaborative interaction characteristics that would likely provide valuable additional information when attempting to characterize the roles. For instance, the affective characteristics of individuals and groups have been shown to play a very important role in learning (Baker, D'Mello, Rodrigo, & Graesser, 2010; D'Mello & Graesser, 2012; Graesser, D'Mello, & Strain, 2014). There has also been evidence suggesting the importance of micro-behavioral measures, such as keystrokes, click-streams, response time, duration, and reading time measures, that could provide additional information (Antonenko et al., 2012; Azevedo, et al., 2010; Mostow & Beck, 2006). Finally, although we used the measure of topic relevance as an independent measure of group performance (i.e., separate from student learning gains) in the current work, this is arguably a feature that could provide valuable information for understanding social roles in group interactions. These limitations will be addressed in subsequent research.

In the Detecting Social Roles section, the GCA was applied to two large, collaborative learning datasets, and one collaborative problem solving dataset (learner $N = 2,429$; group $N = 3,598$). Participants were then clustered based on their profiles across the GCA measures. The cluster analyses identified roles that have distinct patterns in behavioral engagement style (i.e., active or passive, leading or following), contribution characteristics (i.e., providing new information or echoing given material), and social orientation. The six-cluster model revealed the following roles: ***Drivers, Task-Leaders, Socially Detached learners, Over-riders, Followers, and Lurkers***. The identified social roles (i.e., clusters) underwent stringent evaluation, validation, and generalization assessments. The bootstrapping and replication analyses illustrated that the roles generalize both within and across different collaborative interaction datasets, indicating that these roles are robust constructs across different experimental contexts. Interestingly, the GCA revealed roles that do not entirely overlap with those observed in Strijbos and De Laat's (2010). The identification of these new roles might serve as a conceptual basis for future research focusing on understanding the social roles within multi-party communication.

We investigated the practical value of the of the identified roles, and whether they were meaningfully related to student learning gains and group performance. Overall, the results suggest that the roles that learners occupy influences their learning, and that the presence of specific roles within a group can be either more or less beneficial for the collaborative outcome.

This analysis yielded two important contributions to the collaborative learning literature. Firstly, the multi-level mixed-effects models applied in this chapter are rarely applied in CSCL research; however, they are the most appropriate statistical analysis for this nested structure data CSCL data (De Wever et al., 2007; Janssen et al., 2011; Pinheiro & Bates, 2000). Furthermore, these models impose a very stringent test of the influence of roles on group and individual performance by controlling for the variance associated with each participant and group. As such, the use of mixed-effects models provides confidence in the robustness of the findings. Secondly, the multi-level investigation addressed a frequently noted limitation found in collaborative learning research. As Kapur et al., (2011) wrote:

*"It is worth reiterating that these methods should not be used in isolation, but as part of a larger, multiple grain size analytical program. At each grain size, findings should potentially inform and be informed by findings from analysis at other grain sizes—an analytical approach that is commensurable with the multiple levels (individual, group) at which the phenomenon unfolds. Only then can these methods and measures play an instrumental role in the building and testing of a process-oriented theory of problem solving and learning."*

Some of the most noteworthy discoveries concern the influence of roles on student learning and group performance, which suggest the difference in outcome measures across the social roles is not a result of individuals simply being more prolific. Simply participating a lot is far less important than is the nature of that participation (as captured by the internal cohesion, responsivity, and social impact measures). That is, the quality of conversation, more than the quantity, appears to be the key element in the success for both groups and individuals.

One of the central contributions of the GCA can also be viewed as a limitation. One of the benefits of the preconceived categories involved in manual content analyses is that these coded categories would afford a "gold standard" external validation. For instance, if these roles were identified through manually coded categories, then the cluster analysis results could be compared against the human annotated "gold standard". By pursuing a purely automated computational-linguistic methodology, we were able to explore a substantially larger number of collaborative interactions than could be analyzed with manual methods. Furthermore, given the complex and dynamic nature of the discourse characteristics that are calculated in the construction of the GCA, it would be extremely difficult and time consuming, if not impossible, for human coders to capture such multifaceted discourse characteristics. However, external cluster validation can be achieved either by comparing the cluster solutions to some "gold standard" categories or by comparing them to meaningful external variables (Antonenko et al., 2012). In the current research, we successfully took the latter approach by showing that the identified roles are related to both individual student learning and group performance in general, and that the relationship is theoretically meaningful. Furthermore, even "gold standard" human coding schemes must be validated and tested for robustness. We feel that the tests of cluster stability, coherence and internal consistency that we applied to our model are at least as extensive and rigorous as any inter-rater reliability study of a manual coding schema.

This research serves as an initial investigation with the GCA into understanding why some groups perform better than others. Despite some limitations, this research has provided some fruitful lines of research to be pursued in future work. Most significantly, the GCA provides us with a framework to investigate how roles are constructed and maintained through the dynamic sociocognitive processes within an interaction. Individual participants' patterns of linguistic coordination and cohesion, as measured by the GCA, can diagnostically reveal the roles that individuals play in collaborative discussions. As a methodological contribution therefore, we expect the GCA will provide a more objective, domain independent means for future exploration of roles than has been possible with manual coding rubrics. Moreover, as a practical contribution, substantially larger corpora of data can be analyzed with the GCA than when human time is required to annotate the data. Furthermore, the empirical findings of this research will contribute to our understanding of how individuals learn together as a group and thereby advance the cognitive, social and learning sciences.

**Author Note** This research was supported by the Army Research Institute (W5J9CQ12C0043) and the National Science Foundation (IIS-1344257). Any opinions, findings, and conclusions or recommendations expressed in this material are those of the authors and do not necessarily reflect the views of these funding agencies.

# Group Communication Analysis: A Computational Linguistics Approach for Detecting Sociocognitive Roles in Multi-Party Interactions

## - Supplementary material -

Nia M. M. Dowell, Tristan M. Nixon, and Arthur Graesser

## Construction of Group Communication Analysis (GCA) and Group Performance Measure

### Computing LSA Spaces

Each dataset was used to generate a distinct LSA space used for calculating the GCA measures on that dataset. This ensures that each corpus of chat transcripts is given an appropriate semantic representation for the material being discussed. The principal difficulty in generating an LSA space from chat transcripts is that subjects and topics referenced in natural conversations are not sufficiently defined to provide a comprehensive mapping of their semantic relationships. We take for granted that our conversational partners already have a well-developed understanding of a vast array of topics. For example, one may engage in a perfectly coherent conversation with a friend or colleague about careers, food, family or any number of other subjects, without ever needing to provide a comprehensive verbal description of any of these subjects. Therefore, we must supplement the chat transcripts with appropriate external documents in order to robustly represent the semantic space of subjects referenced in a conversation (Cai, Burkett, Morgan, & Shaffer, 2011). To this end, we analyze the frequencies of terms used in the discussion in order to identify the most significant terms (keywords), and then query publicly available databases (i.e., Wikipedia) for documents on those topics. This process of scanning for keywords can be repeated with the newly added documents until a satisfactory number of documents has been obtained to generate a reasonable mapping of the semantic space. Finally, an LSA space of 300 dimensions was computed from each expanded corpus (as described in Chapter 4, above). Table S1 provides the descriptive information for the original chat corpora, the extended corpora, and LSA spaces for each data set.

Table S1

*Total Terms and Unique Terms for each Data Set, Expanded Corpus, and LSA Space*

| Dataset | Chat Transcripts | | Expanded Corpus | | LSA Space |
|---|---|---|---|---|---|
| | Total Terms | Unique Terms | Total Terms | Unique Terms | Unique Terms |
| Traditional CSCL | 130,946 | 6,010 | 2,703,978 | 91,613 | 32,297 |
| SMOC | 457,639 | 14,207 | 8,024,354 | 149,188 | 56,609 |
| Land Science | 401,652 | 9,932 | 1,981,589 | 73,702 | 25,417 |

**Spanning Window Calibration**

The size of the spanning window, w, can have significant effects on the GCA measures. We want to constrain the size of this window so as to capture the temporal dynamics of the conversation (a window as long as or longer than the entire conversation would just average everything together). However, very short windows may miss salient connections between remarks because they fall outside of the specified span. Certain students were such infrequent participants that small window lengths would make computing the w-spanning internal cohesion measure impossible, as all of their contributions were more than w turns apart. A window size of 20 was chosen as this was the shortest length that would allow for at least 95% of students, across all three datasets, to have at least 2 contributions inside the window. The remaining students (< 5%) had their internal cohesion measures trivially set to 0.

## Methods

The top 10 words for the relevant topics are reported in Table 5.

Table 5
*Top Ten Words Representing Eight Relevant Topics*

| Number | Psychological Disorders | General Psychology | Autism | Anxiety Disorder |
|--------|-------------------------|--------------------|--------|------------------|
| 1 | Experience | Association | Child | Percent |
| 2 | Person | Psychology | Autism | Anxiety |
| 3 | Animal | Test | Syndrome | Treat |
| 4 | Schizophrenia | Journal | Autistic | Occur |
| 5 | Thought | Process | Parent | Fear |
| 6 | Study | Addiction | Movement | Blood |
| 7 | Bipolar | Psychiatry | Developmental | Cell |
| 8 | Disorder | Alcohol | Development | Severe |
| 9 | Mental | OCD | Attachment | Pneumonia |
| 10 | Many | Library | Disability | Infection |

| Number | Trauma | Psychotherapy | Personality Disorder | Health Care |
|--------|--------|---------------|----------------------|-------------|
| 1 | Injury | Psychotherapy | Personality | Health |
| 2 | Loss | Technique | Criterion | Care |
| 3 | Bone | Therapist | Diagnostic | Nurse |
| 4 | Speech | Method | ADH | Hospital |
| 5 | Head | Counseling | Statistical | Physician |
| 6 | Surgery | Gun | Trait | Professional |
| 7 | Sound | Start | Sir | Education |
| 8 | Sign | Round | DSM | National |
| 9 | Transsexual | Intervention | Difference | Doctor |
| 10 | Muscle | Game | DSM-IV | Institute |

## Detecting Social Roles

Density and pairwise scatter plots for the GCA variables are reported in Figure 1.

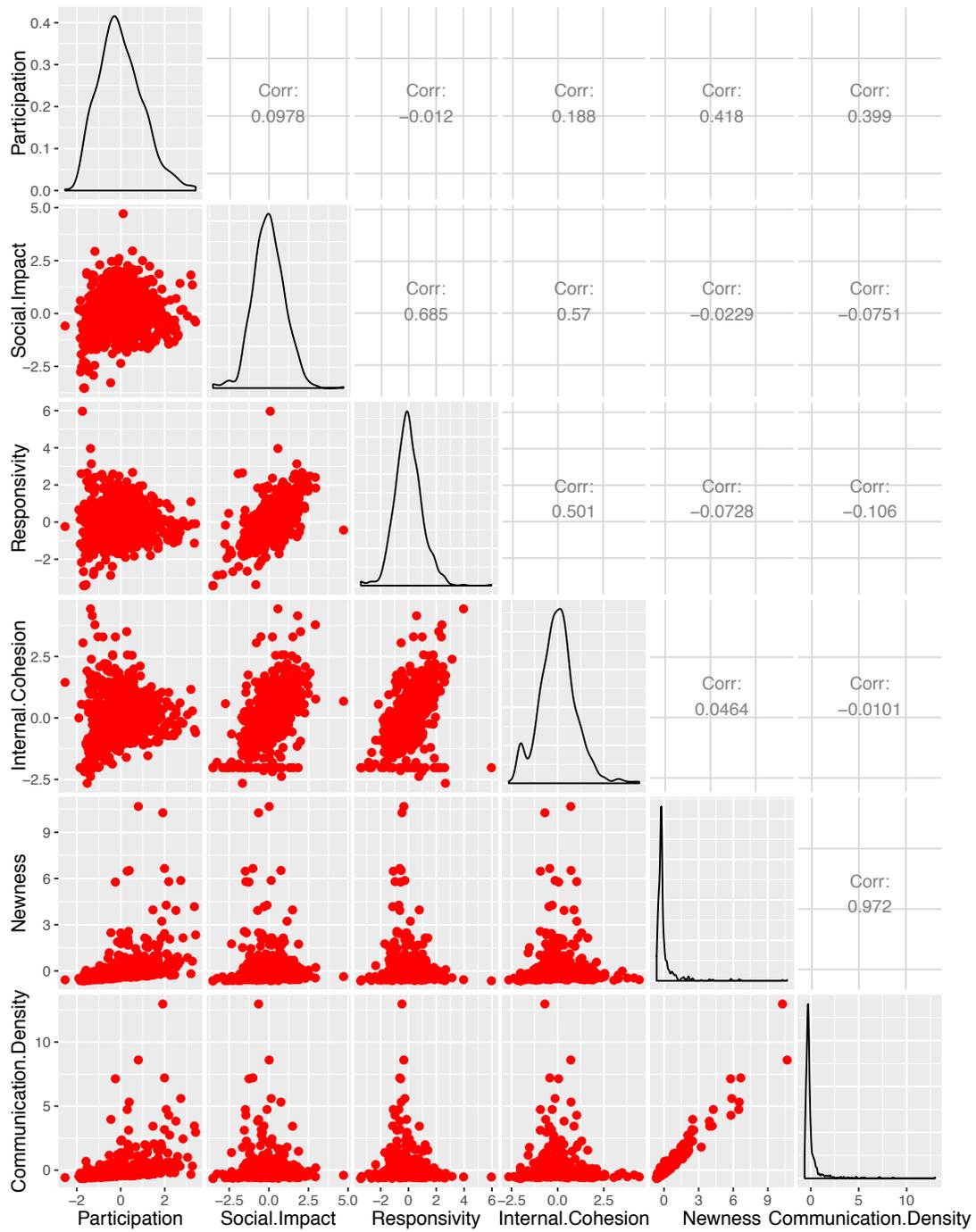

Figure S1. Density and Scatter Plots for GCA Variables

**Collinearity and Multicollinearity Assessment**

Table S2 shows the Pearson correlations between the group communication variables ranged from $r$ = -0.10 to 0.90. The rule-of-thumb is not to use variables correlated at $|r| \geq 0.7$. The VIF values for the group communication variables ranged from 1.65 to 7.34. A rule of thumb states that there is evidence of multicollinearity if VIF > 10 (Fox & Weisberg, 2010). The VIF results support the view that multicollinearity was not an issue. However, there was evidence of moderate collinearity between two variables, newness and communication density. Therefore, the impact of collinearity on the cluster patterns is evaluated further.

Table S2

*Pearson Correlations Coefficients for GCA*

| Measure | Participation | Social Impact | Responsivity | Internal Cohesion | Newness |
|---|---|---|---|---|---|
| Social Impact | 0.07 | | | | |
| Overall Responsivity | -0.01 | 0.69*** | | | |
| Internal Cohesion | 0.21*** | 0.57*** | 0.52*** | | |
| Newness | 0.64*** | 0.07 | -0.03 | 0.10** | |
| Communication Density | 0.56*** | -0.10*** | -0.19*** | -0.06 | 0.90*** |

Note: *** $p$ < .001, ** $p$ < .01, * $p$ < .05.

The potential harm of collinearity in cluster analysis is that is can change the observed pattern of the clusters. The impact of collinearity was evaluated in the current research by running the cluster analyses with and without the communication density measure to ensure the same cluster pattern was observed. This evaluation showed that collinearity was not impacting the cluster pattern for the six-cluster model. Specifically, Figure S2 shows the same cluster pattern was observed across the variables when the cluster analysis was conducted without the communication density measure.

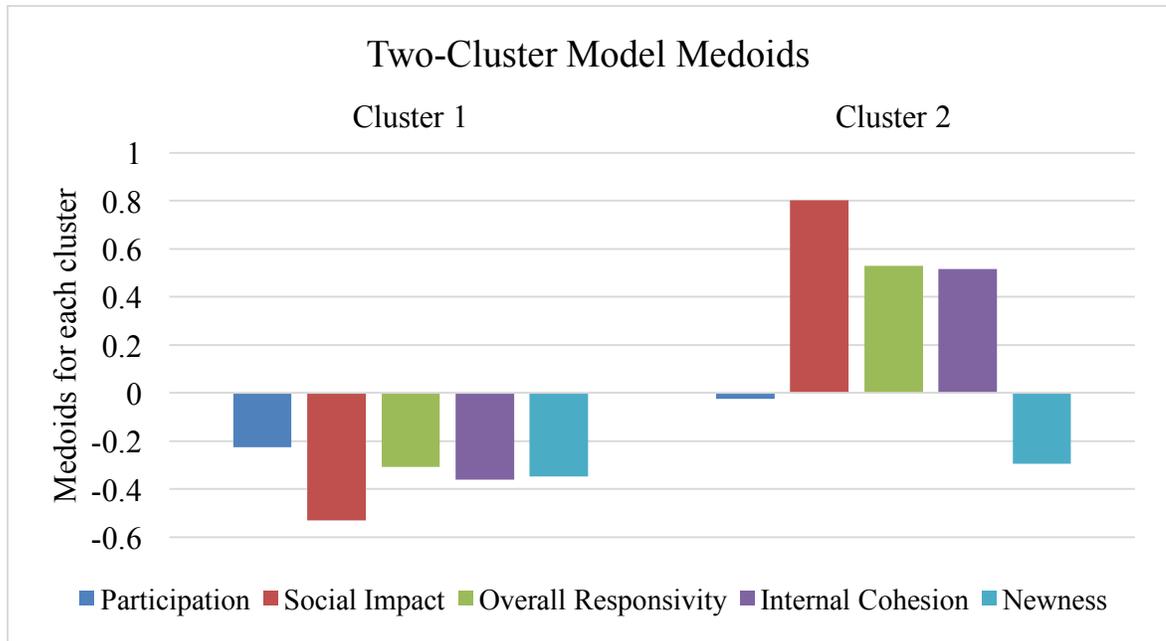

Figure S2. Medoids for the Two-Cluster Solution Without the Communication Density Measure Across the GCA Variables

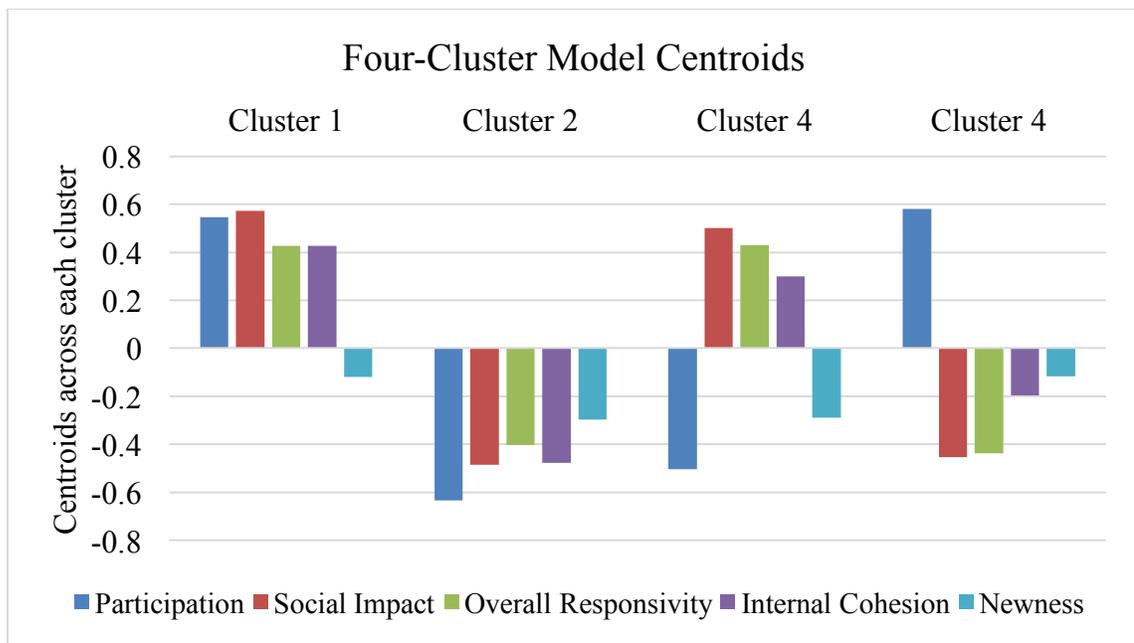

Figure S3. Centroids for the Four-Cluster Solution Without the Communication Density Measure Across the GCA Variables

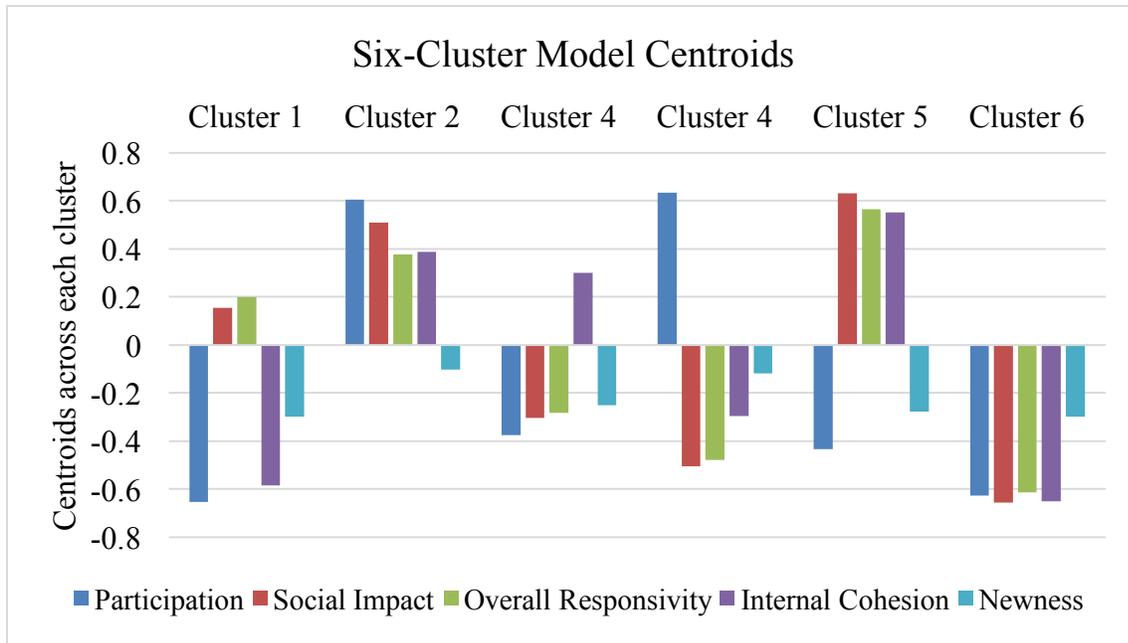

Figure S4. Centroids for the Six-Cluster Solution Without the Communication Density Measure Across the GCA Variables

**Cluster Tendency**

The first step in the clustering process is to assess the cluster tendency (Han, Pei, & Kamber, 2011). Cluster tendency assessment determines whether a given dataset has a non-random structure, which may lead to meaningful clusters. This is a particularly important in the context of unsupervised machine learning because clustering methods will return clusters even if the data does not contain any inherent clusters. The Hopkins statistic is most common method for testing the intrinsic ability of a data to be clustered (Han et al., 2011). The Hopkins statistic is a spatial statistic that tests the spatial randomness of data as distributed in space. The values of the Hopkins statistic (H) ranges from 0 to 1. It tests the null hypothesis that the data are uniformly distributed and thus contains no meaningful clusters. When a dataset is random, implying a lack of underlying structure, the value of H is about .5 or greater. However, when the data exhibit some inherent clustering the H is closer to 0 (Han et. al., 2011, p. 486). In the current project, the Hopkins statistic was implemented, using the R library clustertend (YiLan & RuTong, 2015), to evaluate the cluster tendency for the Traditional CSCL data set prior to conducting the actual cluster analyses. A random uniform simulated dataset was generated with the same dimension as the Traditional CSCL dataset to serve as an illustrative baseline comparison. As expected the random dataset did not exhibit any meaningful clusters, H = .51. However, the Traditional CSCL dataset did show evidence of clustering, H = .11, which is well below the threshold of H > .5.

**Determining Optimal Number of Clusters**

In the current research, both visual approaches such as the 'Elbow' method, and a group of other statistical approaches were explored. The Elbow method is a useful visual way to choose the appropriate number of clusters. The Elbow method involves plotting the wss against a series of sequential cluster levels. The most appropriate cluster solution is defined as the solution at which the reduction in wss slows considerably. This produces an "elbow" in the plot of wss against cluster solutions. To identify the appropriate number of clusters in the Traditional CSCL data set the wss was compared with the number of clusters ranging from 1 to 10. By plotting the number of clusters against the within-groups sum of squares for the group communication variables (Figure 4) it is possible to not only quantitatively, but also visually identify a representative number of clusters. Figure 4 shows that similar values of the within-groups sum of squares appear for values of k greater than four, therefore indicating that four seems to be an appropriate value for the number of clusters to consider. This is in line with on Strijbos and De Laat (2010) conceptual model of student roles.

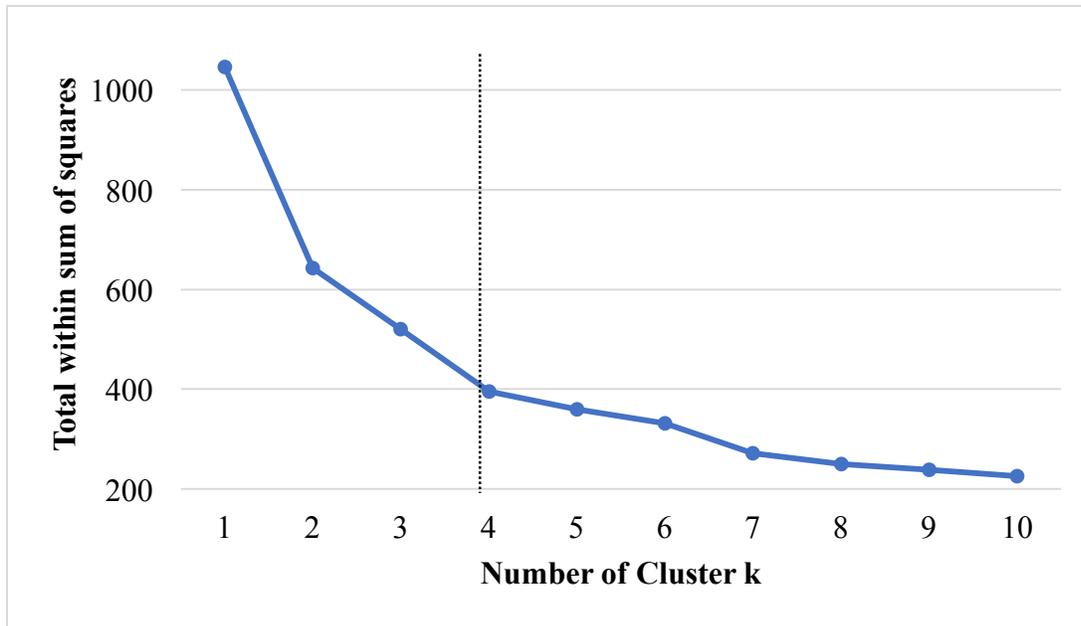

*Figure 4*. Number of clusters solutions against within-groups sum of squares for Traditional CSCL data set GCA variables. Here we see the proposed number of clusters is 4.

The disadvantage of elbow and similar methods (i.e., average silhouette method) is that they provide only a visual impression of clustering without quantitatively measuring the inflection point of the elbow. As mentioned earlier, several indices have been proposed in the literature for determining the optimal number of clusters (Han et al., 2011). Thus, a more precise and comprehensive evaluation would involve exploring the best clustering scheme from the different results obtained by varying all combinations of number of clusters, distance measures (e.g., Manhattan distance for k-medoids, Euclidean distances for k-centroids) and clustering methods. The NbClust package provides 26 indices for determining the relevant number of clusters (Charrad, Ghazzali, Boiteau, & Niknafs, 2014). It is beyond the scope of this project to specify each index, but they are described comprehensively in the original paper of Charrad et al. (2014). An important advantage of NbClust is that researchers can simultaneously compute multiple indices and determine the number of clusters using a majority rule. The majority rule is based on the evaluation of the cluster size proposed across the 26 indices with the final suggested number of clusters based on the majority. In the current project, the optimal number of clusters was explored for two clustering partitioning approaches, Partitioning Around Medoids (PAM) and Partitioning Around Centroids (K-means). Figures 5 and 6 reveal that the optimal number of clusters, according to the majority rule, is 2 for the PAM approach and 6 for the K-means approach. However, the total within-cluster sum of squares (wss) suggested a four-cluster solution. Based on this discrepancy, three models (i.e., the two-, four- and six-cluster solutions) were constructed and compared.

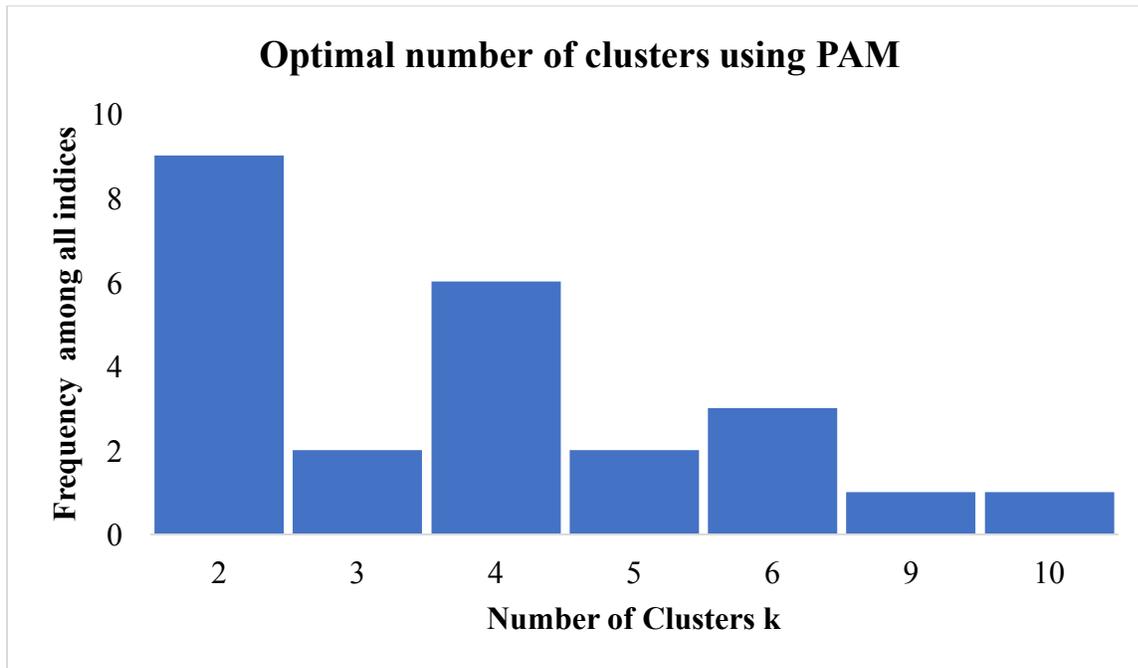

*Figure 5.* Frequency for recommended number of clusters using PAM, ranging from 2 to 10, using 26 criteria provided by the NbClust package. Here we see 9 of the 26 indices proposed 2 as the optimal number of clusters in the Traditional CSCL dataset.

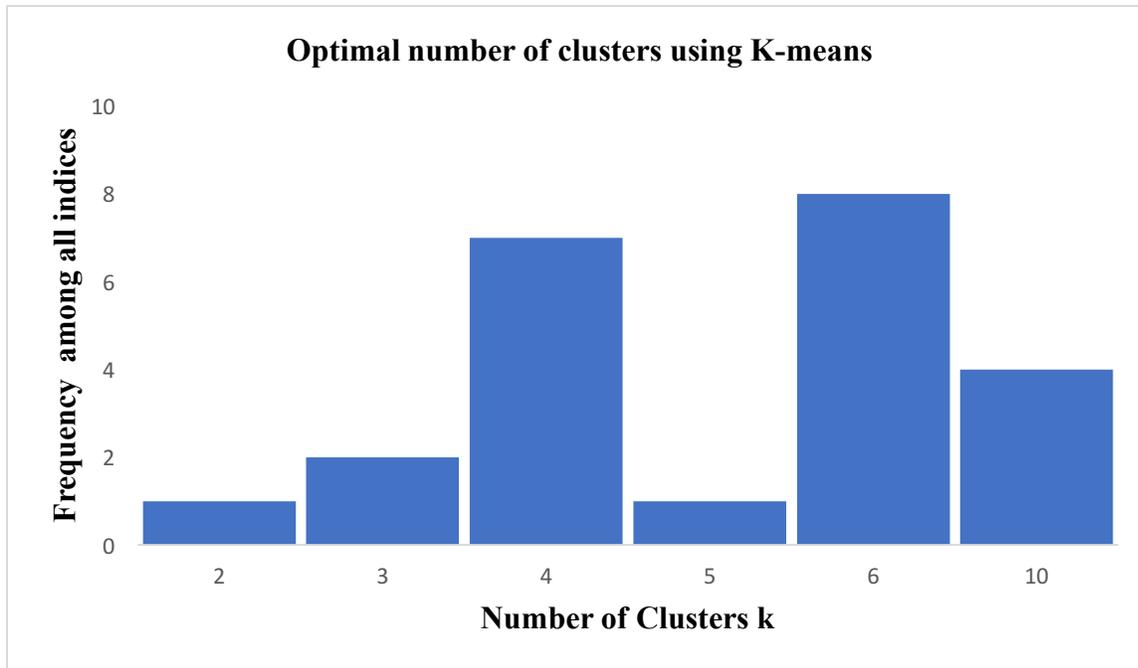

*Figure 6.* Frequency for recommended number of clusters using K-means, ranging from 2 to 10, using 26 criteria provided by the NbClust package. Here we see 8 of the 26 indices proposed 6 as the optimal number of clusters in the Traditional CSCL dataset.

**Partitioning Clustering Analysis (Unsupervised Analysis)**

Partitioning based clustering methods include two major categories, namely k-means and k-medoids. While several partitioning methods were explored in the current dissertation (including PAM, fuzzy, hierarchical, density, hybrid k-means and regular k-means clustering), PAM and k-means provided the most stable clusters. Thus, the PAM and k-means methods were used to group learners with similar group communication profiles into clusters. Three separate cluster analyses were performed to assess the degree to which the data resembled a two-, four- or six-cluster solution. A first step in interpreting the clusters involves inspecting the cluster centroids for k-means, or medoids for PAM, as this sheds light on whether the segments are conceptually distinguishable. Centroids are representative objects, or in this context learners, of a cluster whose average dissimilarity to all the other learners in the cluster is minimal. Centroids are conceptually similar to means. In contrast to the centroids used in the $k$-means algorithm, the medoids from PAM are represented by actual data points that best characterize the cluster. The medoids for the two cluster PAM solution, and centroids for the four- and six-cluster k-means solution are presented below in Figures 7-9, respectively.

 As discussed earlier, there was evidence of moderate collinearity between two variables, newness and communication density. The potential harm of collinearity in cluster analysis is that is can change the observed pattern of the clusters. The impact of collinearity was evaluated in the current research by running the cluster analyses with and without the communication density measure to ensure the same cluster pattern was observed. This evaluation showed that collinearity was not impacting the cluster pattern for the two-, four-, or six-cluster models. Specifically, the same cluster pattern was observed across the variables when the cluster analysis was conducted without the communication density measure (see Appendix D, Appendix E, and Appendix F).

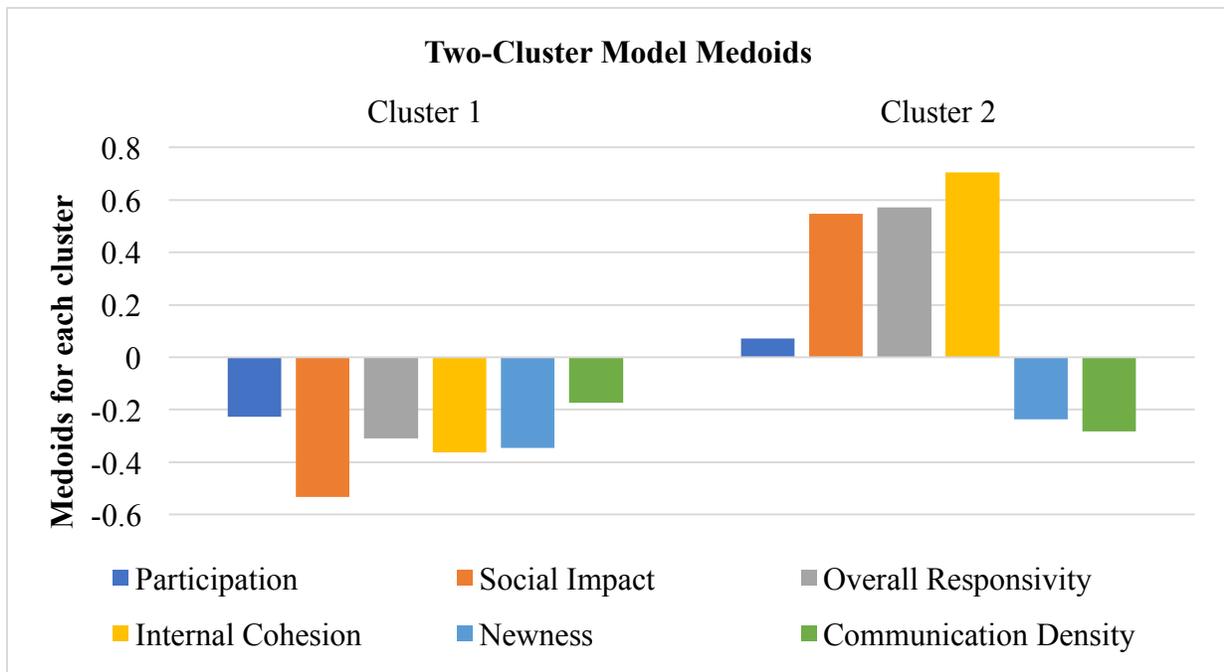

*Figure 7*. Medoids for the two-cluster solution across the GCA variables.

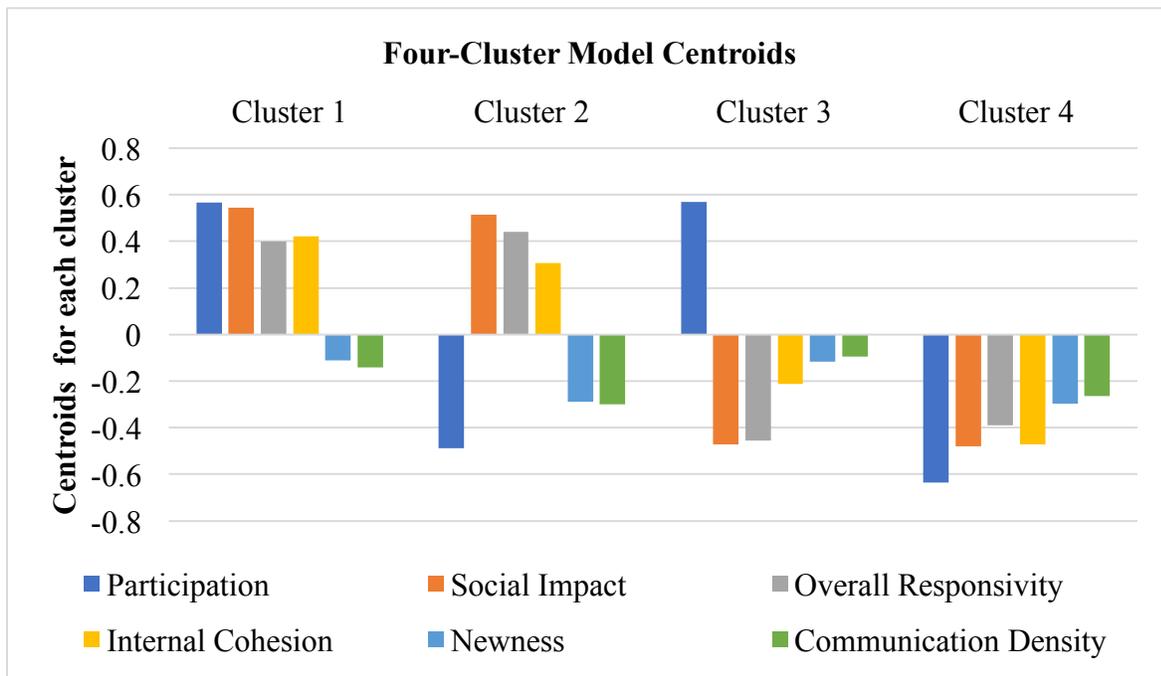

*Figure 8*. Centroids for the four-cluster solution across the GCA variables.

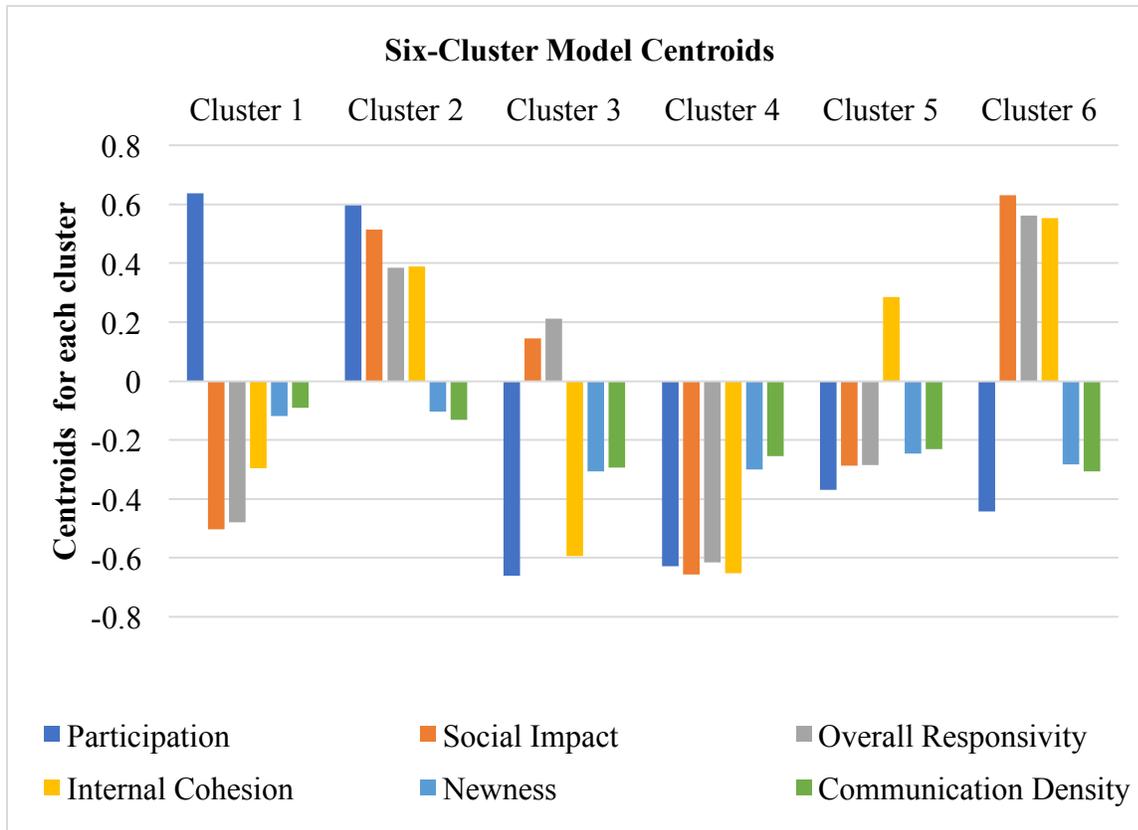

*Figure 9*. Centroids for the six-cluster solution across the GCA variables.

We see some similar patterns across the two-, four-, and six-cluster solutions, which suggest stability in the cluster analysis. Figure 7 shows the two-cluster solution segmented learners who did not productively engage in the collaborative interaction (cluster 1) from those that did (cluster 2). Similar clusters where identified in the four-cluster model (see Figure 8 clusters 1 and 4), and the six-cluster model (see Figure 9 clusters two and four). Specifically, cluster 1 in the two-cluster model (Figure 7), 4 on the four-cluster model (Figure 8), and cluster 4 in the six-cluster model (Figure 9) are characterized by the lowest participation, social impact, overall responsivity, internal cohesion, newness, and communication density. This patterns resembles the Ghost in the Strijbos-DeLaat role framework. Although, that model used the category label Ghost, hereafter in this dissertation it will be labeled the ***Lurker*** role. Lurkers have been defined differently in the literature, ranging from non-participators to minimal participators (Nonnecke & Preece, 2000; Preece, Nonnecke, & Andrews, 2004). The distinction between a Ghost and a Lurker is not clear in the literature, which appears to use these terms interchangeably, although Strijbos and De Laat do make a distinction based on group size. Two reasons motivated operationalizing this pattern as a Lurker, rather than the Ghost, in the current research; First, the GCA methodology would not be able to detect an individual that did not participate at all (because there would not be a log file for those students), which suggests the learners in these clusters did contribute at least minimally. Second, past research has labeled the Ghost and Lurker roles predominantly based on the amount of contributions a student makes, although the GCA captures participation as well as the sociocognitive characteristics of those contributions. The pattern depicted for these clusters does not suggest these students have no social impact, or were completely unresponsive to others. Rather it suggests that these students expressed less compared to other group members. Lurking behavior sometimes involves some level of engagement but at other times little engagement so it is associated with both positive and negative outcomes in the literature (Preece et al., 2004). Therefore, Lurker appeared to be the most appropriate label for this cluster.

Similar patterns were also observed between cluster 2 in the two-cluster model, cluster 1 in the four-cluster model, and cluster 2 in the six-cluster model. The learners in these clusters are among the highest participators; they exhibit high social impact, responsiveness, and internal cohesion, but coupled with the lowest newness and communication density. Learners in these clusters are investing a high degree of effort in the collaborative discussion and display self-regulatory and social-regulatory skills. This pattern is labeled the ***Driver*** in the current research.

While the two-cluster model makes sense conceptually, the simplicity of the segmentation is less meaningful from a practical and theoretical standpoint.

The four and six cluster solutions provide more detail by further distinguishing the mid-range students. For instance, cluster 3 in the four-cluster model and cluster 1 in the six-cluster model are characterized by learners who have the highest participation. However, when they contribute, their discourse is more in response to themselves than other team members (i.e., higher internal cohesion than responsiveness or social impact), and do not warrant further discussion from the group members or provide new information (i.e., low social impact and newness). These individuals would be similar to the Over-riders described in Strijbos and De Laats' (2010) framework, who exhibit strong individual learning goals and try to push the group members into adopting their agenda. In contrast to the Driver role, *Over-riders* have a higher degree of internal cohesion compared to social impact or responsiveness, which signals the Over-rider is more concerned with the personal gain than the collaboration or social climate.

Cluster 2 in the four-cluster model and cluster 6 in the six-cluster model are also quite similar. Here we see learners with low participation, but when they do contribute, they attend to other learners' contributions and provide meaningful information that furthers the discussion (i.e., high internal cohesion, overall responsiveness, and social impact). This pattern is similar to a student that is engaged in the collaborative interaction and is called a *Task-Leader* role in this research. It is interesting to note that these students are not among the highest participators, but their discourse signals a social positioning that is conducive to a productive exchange within the collaborative interaction.

Cluster 3 and 5 in the six-cluster model (Figure 9) produced two additional patterns not observed in the other cluster models. Learners occupying cluster 5 exhibited high internal cohesion, but low scores on all the other group communication measures. This cluster is labeled as ***Social Detached***, because the pattern appears to capture students who are not productively engaged with their collaborative peers, but instead focused on themselves. Cluster 3 is characterized by learners who have the lowest participation. However, when they do contribute it appears to build, at least minimally, on previously contributed ideas and move the collaborative discourse forward (i.e., higher social impact and responsiveness). This cluster is labeled as the ***Follower***. Overall, all three cluster models appear, at least visually, to produce theoretically meaningful student groupings. In the next phase of the analysis the quality and validity of the cluster solutions is evaluated.

**Clustering Evaluation and Validation**

The literature proposes several cluster validation indexes that quantify the quality of a clustering (Hennig, Meila, Murtagh, & Rocci, 2015). In principle, these measures provide a fair comparison of clustering and aid researchers in determining whether a particular clustering of the data is better than an alternative clustering (Taniar, 2006). There are three main types of cluster validation measures and approaches available: internal, stability, and external. Internal criteria evaluate the extent to which the clustering "fits" the data set based on the actual data used for clustering. In the current dissertation three commonly reported internal validity measures (Silhouette, Dunns index, and Connectivity) were explored using the R package clValid (Brock, Pihur, Datta, & Datta, 2008). Silhouette analysis measures how well an observation is clustered and it estimates the average distance between clusters (Rousseeuw, 1987). Silhouette widths indicate how discriminant the clusters chosen are by providing values that range from -1, indicating that observations are likely placed in the wrong cluster to 1, indicating clusters perfectly separate the data and no better (competing) ways to cluster can be found. Dunn's index (*D*) evaluates the quality of clusters by computing a ratio between the inter-cluster distance (i.e., between cluster separation) and intra-cluster diameter (i.e., within-cluster compactness). Larger values of D suggest good clusters, and a D larger than 1 indicates compact separated clusters (Dunn, 1974). The Connectivity measure captures the extent to which observations are placed in the same cluster as their nearest neighbors (Handl, Knowles, & Kell, 2005). The connectivity has a value between zero and $\infty$ and should be minimized. These internal stability measures for the two-, four-, and six-cluster solutions are reported in Table 13. As can be seen from Table 13, the two-cluster solution had the highest internal validity across the three measures, followed by the four-cluster solution. The two-cluster model was substantially better for the Connectivity measure. However, for the Dunn Index and Silhouette measures, the two-cluster model was only marginally better than the other cluster solutions. For instance, we see the two-cluster solution, compared to the four-cluster solution, is only .2 higher for the Silhouette measure, and .01 higher for the Dunn Index.

Table 13

*Internal Validity Measures for the Two, Four, and Six Cluster Solutions*

| Internal Validity Measures | Two-Cluster Model | Four-Cluster Model | Six-Cluster Model | Index Range/ Preference |
|---|---|---|---|---|
| Silhouette | .33 | .30 | .31 | Zero to one/ Higher |
| Dunn Index | .07 | .06 | .05 | Zero to $\infty$/ Higher |
| Connectivity | 87.72 | 196.01 | 249.55 | Zero to $\infty$/ Lower |

Stability is another important aspect of cluster validity. Stability means that a meaningful valid cluster should remain intact (i.e., not disappear easily) if the data set is changed in a non-essential way (Hennig, 2007). While there are different conceptions of what constitutes a "non-essential change" of a data set, a common method employed is the leave-one-column out. The stability measures compare the results from clustering based on the full data set to clustering based on removing each column, one at a time (Brock et al., 2008; Datta & Datta, 2003). In the current data set this corresponds to the removal of one of the GCA variables at a time. The stability measures are the average proportion of non-overlap (APN), the average distance (AD), the average distance between means (ADM), and the figure of merit (FOM). The APN measures the average proportion of observations not placed in the same cluster by clustering based on the full data and clustering based on the data with a single column removed. The AD measure computes the average distance between observations placed in the same cluster by clustering based on the full data and clustering based on the data with a single column removal. The ADM measure computes the average distance between cluster centers for observations placed in the same cluster by clustering based on the full data and clustering based on the data with a single column removed. The FOM measures the average intra-cluster variance of the observations in the deleted column, where the clustering is based on the remaining (undeleted) samples. This estimates the mean error using predictions based on the cluster averages (Brock et al., 2008). In all cases the average is taken over all the deleted columns, and all measures should be minimized. As seen in Table 14, the stability scores for the two-, four-, and six-cluster solutions are quite similar, with the two-, and four-cluster solution being, on average, only slightly more stable than the six-cluster model. The results from the internal validity and stability inspection showed, on average, only minimal differences between the cluster solutions. However, the two-cluster solution only categorized learners as high and low across the GCA variables (see Figure 7). This simple dichotomous grouping is less meaningful for identifying more intricate conversational patterns of students' social roles. Therefore, the four-cluster and six-cluster solutions were chosen in moving forward. In subsequent analyses.

Table 14

*Stability Validity Measures for the Two, Four, and Six Cluster Solutions*

| Stability Measures | Two-Cluster Model | Four-Cluster Model | Six-Cluster Model | Index Range |
|---|---|---|---|---|
| Average proportion of non-overlap (APN) | .14 | .18 | .22 | Zero to one |
| Average Distance (AD) | 1.31 | 1.07 | .97 | Zero to $\infty$ |
| Average Distance between means (ADM) | .23 | .26 | .31 | Zero to $\infty$ |
| Figure of merit (FOM) | .40 | .38 | .37 | Zero to $\infty$ |

**Cluster Coherence**

It is important to evaluate the coherence of the clusters from a statistical analysis of the GCA variables involved in their partitioning. Consequently, the four- and six-cluster models were further evaluated to determine whether learners in the cluster groups significantly differed from each other on the six GCA variables. The multivariate skewness and kurtosis were investigated using the R package MVN (Korkmaz, Goksuluk, & Zararsiz, 2015) which produced the chi-square Q-Q plot (see Figure S3) and a test statistic Henze-Zirkler (*HZ*) which assesses whether the dataset follows an expected multivariate normal distribution. The results indicated the GCA variables did not follow a normal distribution, $HZ = 5.06$, $p < .05$. Therefore, a permutational MANOVA (or nonparametric MANOVA) was used to test the effect of the four and six-cluster models on the GCA variables.

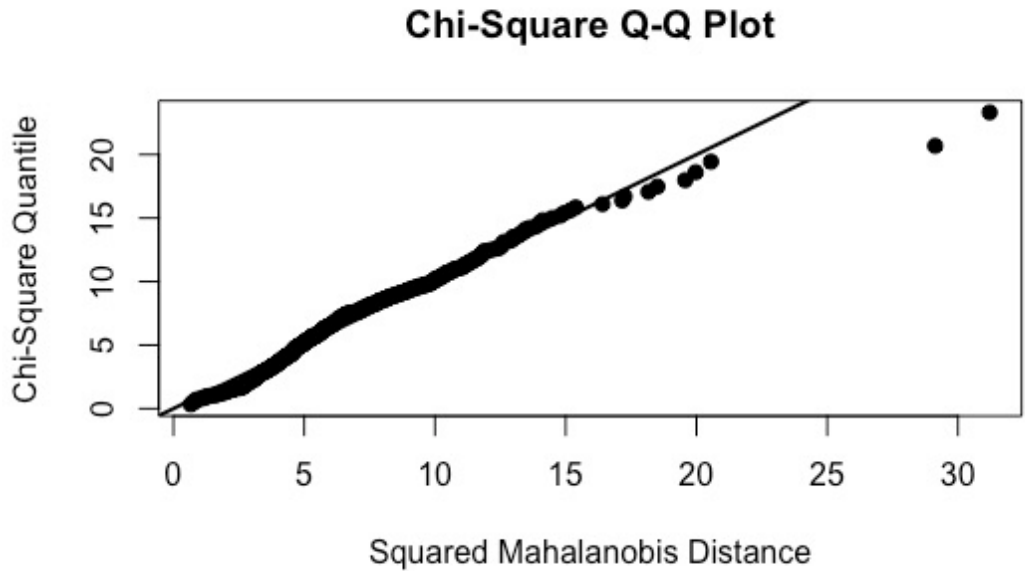

Figure S3. Multivariate skewness and kurtosis evaluation chi-square Q-Q plot

The permutational MANOVA, implemented in the *Adonis* routine of the VEGAN package in R

(Oksanen et al., 2016), is a robust alternative to both parametric MANOVA and to ordination

methods for describing how variation is attributed to different experimental treatments or, in this

case, cluster partitions (Anderson, 2001). The *Adonis* test showed a significant main effect of

cluster for the four-cluster model, $F(3,714)=392.21$, $p < .001$, and six-cluster model, $F$

$(5,712)=350.86$, $p < .001$. These results support the models' formation and ability to organize

learners based on differences in their collaborative communication profiles.

The analyses proceeded with ANOVAs followed by Tukey's post hoc comparisons to characterize learners' patterns by identifying significant differences in participants' scores on the six GCA variables between the clusters. Levene's Test of Equality of Error Variances was violated for all the GCA variables so a more stringent alpha level ($p < .01$) was used when identifying significant differences for these variables (Tabachnick & Fidell, 2007, p. 86). The ANOVA main effect $F$-values along with the means and standard deviations for the GCA variables across each cluster are reported in Table 15 for the four-cluster model, and Table 16 for the six-cluster model. The ANOVA revealed significant differences among clusters for all of the six GCA variables at the $p < .0001$ level for both the four and six-cluster models. Tukey's HSD post hoc comparisons for the four and six-cluster models are presented in Table 17 and Table 18, respectively. As seen in Table 17 and Table 18, the post hoc comparisons confirmed that the observed differences in GCA profiles across the clusters were, for the majority, significantly distinct in both models. In the next phase of the analysis, the four and six-cluster models were further examined to determine external validity

Table 15

*Four-cluster Model Means and Standard Deviations for the 6 GCA Variables*

| GCA Measures | Cluster 1: Driver $n$ = 154 $M(SD)$ | Cluster 2: Task-Leader $n$ = 182 $M(SD)$ | Cluster 3: Over-rider $n$ = 171 $M(SD)$ | Cluster 4: Lurker $n$ = 211 $M(SD)$ | *F*-value |
|---|---|---|---|---|---|
| Participation | 0.57(0.26) | -0.49(0.3) | 0.57(0.29) | -0.64(0.27) | 440.30*** |
| Social Impact | 0.55(0.3) | 0.52(0.35) | -0.47(0.31) | -0.48(0.38) | 282.70*** |
| Overall Responsivity | 0.40(0.39) | 0.44(0.37) | -0.45(0.32) | -0.39(0.44) | 173.80*** |
| Internal Cohesion | 0.42(0.31) | 0.31(0.47) | -0.21(0.41) | -0.47(0.41) | 130.90*** |
| Newness | -0.11(0.14) | -0.29(0.13) | -0.12(0.14) | -0.3(0.14) | 27.09*** |
| Communication Density | -0.14(0.16) | -0.3(0.13) | -0.1(0.14) | -0.26(0.15) | 25.06*** |

Note: ANOVA *df* = 3,714; *** *p* < .0001

**Model Generalizability**

**Internal generalizability**

Descriptive statistics for the GCA variables are reported below in Table S3.

Table S3

*Descriptive Statistics for GCA Measures in the Traditional CSCL Testing Data Set*

| Measure | Minimum | Median | *M* | *SD* | Maximum |
|---|---|---|---|---|---|
| Participation | -0.23 | -0.01 | 0.00 | 0.10 | 0.30 |
| Social Impact | -0.01 | 0.18 | 0.18 | 0.05 | 0.33 |
| Overall Responsivity | 0.00 | 0.18 | 0.18 | 0.05 | 0.41 |
| Internal Cohesion | 0.00 | 0.20 | 0.19 | 0.11 | 1.00 |
| Newness | 0.05 | 0.49 | 0.72 | 1.06 | 11.04 |
| Communication Density | 0.01 | 0.21 | 0.32 | 0.49 | 5.23 |

*Note*. Mean (***M***); Standard deviation (***SD***); $N = 136$.